\documentclass[10pt,journal, final, twocolumn]{IEEEtran}

\usepackage{epsfig}
\usepackage{graphicx}
\usepackage{amsmath}
\usepackage{amssymb}
\usepackage{multirow}
\usepackage{rotating}
\usepackage{subfigure}
\usepackage{bm}
\usepackage{color}

\begin{document}


\title{Improving Texture Categorization with Biologically Inspired Filtering}

%

\author{Ngoc-Son Vu, Thanh Phuong Nguyen, Christophe Garcia
\thanks{Ngoc Son Vu (son.vu@ensea.fr) is with the ETIS lab - ENSEA, Univ. Cergy Pontoise, CNRS. Thanh Phuong Nguyen is with ENSTA ParisTech, and Christophe Garcia is with the LIRIS lab, INSA Lyon.}
}

\markboth{}{Vu et al.: Improving Texture Categorization with Biologically
Inspired Filtering}

\maketitle

\begin{abstract}

Within the domain of texture classification, a lot of effort has been spent on local descriptors,
leading to many powerful algorithms. However, preprocessing techniques have received much less attention
despite their important potential for improving the overall classification performance. We address
this question by proposing a novel, simple, yet very powerful biologically-inspired filtering
(BF) which simulates the performance of human retina. In the proposed approach, given a
texture image, after applying a DoG filter to detect the ``edges'', we first split the
filtered image into two ``maps'' alongside the sides of its edges. The feature extraction
step is then carried out on the two ``maps'' instead of the input image. Our algorithm has
several advantages such as simplicity, robustness to illumination and noise, and
discriminative power. Experimental results on three large texture databases show that with
an extremely low computational cost, the proposed method improves significantly the
performance of many texture classification systems, notably in noisy environments. The source codes of the proposed algorithm can be downloaded from https://sites.google.com/site/nsonvu/code.
\end{abstract}



\section{Introduction}
Texture classification is a fundamental issue in computer vision and image processing, playing
a significant role in many applications such as medical image analysis, remote
sensing, object recognition, document analysis, environment modeling, content-based
image retrieval and many more. As the demand of such applications increases, texture
classification has received considerable attention over the last decades and numerous
novel methods have been proposed \cite{cula2001btf,ojala2002pami,lazebnik2005pami,varma2005ijcv,varma2009pami,liao2009tip,clbp2010tip,chen2010weber, bif2010,liu2012imavis,clbc2012tip,bgp2012,timofte2012bmvc}.



The texture classification problem is typically divided into the two subproblems of representation and
classification \cite{randen1999pami}, and to improve the overall quality of texture classification, researchers
often focus on improving one of (or both) those steps. It is generally agreed that texture features play a very
important role, and the last decade has seen numerous powerful descriptors being proposed such as modified
SIFT (scale invariant feature transform) and intensity domain SPIN images \cite{lazebnik2005pami}, MR8 \cite{varma2009pami}, the rotation
invariant basic image features (BIF) \cite{bif2010}, (sorted) random projections over small patches \cite{liu2012pr}, local binary pattern (LBP) \cite{ojala2002pami}, and
its variants \cite{liao2009tip,clbp2010tip,liu2012imavis,clbc2012tip}.
Also, different similarity measures such as $\chi^2$ statistic \cite{ojala2002pami,varma2009pami},
Bhattarcharyya distance \cite{bif2010}, and Earth Mover's Distance \cite{lazebnik2005pami} are
often used in conjunction with nearest neighbor classifiers \cite{ojala2002pami}, non-linear (kernel-based)
support vector machines (SVMs) \cite{caputo2010}, or collaborative representation-based
classifier \cite{timofte2012bmvc}. Undoubtedly, an efficient preprocessing which enhances
the robustness and discriminative power of texture features is an important factor towards
enhancing such texture classification systems. However, to the best of our knowledge,
there does not exist any ``sufficiently efficient'' preprocessing methods which can significantly
improve texture features. 

\section{Related work}
Most of earlier work on texture analysis focused on the development of
filter banks and on characterizing the statistical distributions of their responses,
e.g., \cite{cula2001btf}, until Ojala et al. \cite{ojala2002pami} proposed LBP
(Local Binary Patterns) and showed that statistics of small pixel neighborhoods
are capable of achieving high discrimination.
Since then, due to its impressive computational efficiency and good texture
characterization, the dense LBP descriptor has gained considerable attention
and has been intensively used in different applications such as face recognition
\cite{ahonen2006pami}.

LBPs are ``micro'' features capturing the distribution of the relationships
between pixels in small-scale neighborhoods, but unfortunately, has several
limitations such as small spatial support region, loss of local textural
information, and rotation and noise sensitivities. To overcome those, a lot
of effort has been done. To recover from the loss of information, local image
contrast was introduced by Ojala et al. \cite{ojala2002pami} as a complementary
measure, and better performance has been reported. By a ``completed''
LBP model, Guo et al. \cite{clbp2010tip} included both the magnitudes of local
differences and the pixel intensity itself, and claimed better performance.
In terms of locality, \cite{liao2009tip} proposed to extract global features
from the Gabor filter responses as a complementary descriptor. Dominant LBP (DLBP)
also presented in \cite{liao2009tip} rely on dominant patterns which were experimentally
chosen from all rotation invariant patterns. Regarding noise robustness,
Ojala et al. \cite{ojala2002pami} introduced the concept of uniform and rotation
invariant patterns (LBP$^{riu2}$), while Tan and Triggs \cite{tan2007a} proposed
local ternary patterns (LTP). Liu et al. \cite{liu2012imavis} have recently
generalized LBP with two different and complementary types of features which are
extracted from local patches, based on pixel intensities and differences.
Using this approach, the authors reported impressive texture classification rates.
In \cite{clbc2012tip}, a ``LBP like'' feature, the Local Binary Count (LBC),
is proposed, in which a pixel is encoded by the number of neighbors whose intensities are
larger than that of the considered pixel. Heikkila et al. \cite{heikkila2009} exploit
circular symmetric LBP (CS-LBP) for local interest region description. Also, presented in \cite{zhao2012tip} are several LBP variants for image and video description, that is, the LBP histogram Fourier (LBP-HF) features, and
the LBPs from three orthogonal planes (LBP-TOP) features. In \cite{chen2010weber}, Chen et al. proposed WLD (Weber Local Descriptor) method  based on the fact that human perception of a pattern depends not only on the change of a stimulus (such as sound, lighting) but also on the original intensity of the stimulus. WLD consists of two components: differential excitation and orientation.

An alternative method to improve the strength of texture descriptor is to perform efficient preprocessing.
For example, in face recognition, Vu and Caplier \cite{vu2010poem,vu2012tip} applied the LBP operators upon three edge distribution maps across different directions, and reported state-of-the-art performance.
However, to the best of our knowledge, with regard to feature extraction in
texture classification, no such efficient preprocessing method exists (in \cite{liao2009tip},
the DLBP features and Gabor filters are extracted separately). This is the motivation for
the algorithm presented in this paper.



Neuroscience has made lots of progress in understanding the visual system and how images are
transmitted to the brain. It is believed that the Difference of Gaussians (DoG) filter simulates
how the human retina processes the images observed and extracts theirs details. We propose to somehow
mimic the same strategy to generate richer and more robust information from the image before
carrying out the feature extraction step.

The rest of the paper is structured as follows. Section ~\ref{ss:algo} details the proposed approach.
Experimental results are presented in Section ~\ref{ss:result} and conclusions are finally drawn
in Section ~\ref{ss:conclusion}.

\section{Proposed method}\label{ss:algo}
The main objective of the proposed method is to enhance robustness and discriminative power of
texture classification systems at the level of preprocessing. To be invariant
to rotation, one of crucial requirements in texture classification, techniques used should
either discard all orientation information or capture relative orientation information. We follow
 the first idea and we use a simple yet efficient isotropic DoG filter which simulates the
 performance of human retina. Briefly speaking, given an input texture image, we first use a DoG
 filter to detect its ``edges'' and then split the filtered image into two ``maps'' alongside two
 sides of the detected edges (the term ``edges'' used here refer to the positions where there are
 changes in intensity). Feature extraction algorithms, e.g., the LBP encoding method, are then
 carried out on those resultant ``maps'' to obtain the final texture representation.

This section first briefly describes the human retina, in particular the bipolar cells by
which our algorithm is inspired. We then detail the proposed method and discuss its properties.

\subsection{Model of Retinal Processing}
\begin{figure}
\begin{center}
\includegraphics[width=\columnwidth]{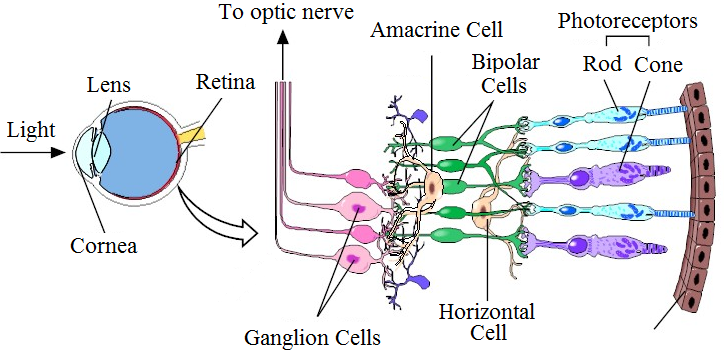}
\end{center}
\caption{The retina lies at the back of the eye. Light passes through the ganglion, amacrine, bipolar, and horizontal
cells and reaches the photoreceptors layers where it returns. Thus the signal is transmitted and processed within the retina in the inverse order, from the photoreceptors, horizontal and bipolar cells
to amacrine and ganglion cells.}
\label{fig:eye}
\end{figure}
The retina lies at the back of the eye (see Fig.~\ref{fig:eye}). Basically, it is made of three
layers: the photoreceptors layer with cones and rods; the outer plexiform layer (OPL) with horizontal,
bipolar and amacrine cells; and the inner plexiform layer (IPL) with ganglion cells. It is worth noting that the retina is capable to process both spatial and temporal signals but working on static images,
we consider only the spatial processing in the retina.

\textbf{Photoreceptors:} rods and cones have quite different properties: rods have the ability to
see at night, under conditions of very low illumination; cones have the ability to deal with
bright signals. Photoreceptor layer plays therefore the role of a light adaptation filter.

\textbf{Outer plexiform layer (OPL):} the photoreceptor performs a low pass filter. Horizontal cells
perform a second low pass filter. In OPL, bipolar cells calculate the difference between photoreceptor
and horizontal cell responses. 
noise and low frequency illumination.

Typically, to model the processes of OPL, two Gaussian low pass filters with different standard
deviations corresponding to the effects of photoreceptors and horizontal cells are used. Thus,
bipolar cells act like a Difference of Gaussians (DoG) filter.

\textbf{Inner plexiform layer (IPL):} IPL works similarly to OPL but it performs on the temporal
information rather than on the spatial one as in OPL.

In the literature, different algorithms inspired by the human retina have been proposed for
different applications \cite{meylan2007,vu2009retina,benoit2005,picot2012retina}. In \cite{meylan2007},
Meylan et al. proposed a tone mapping algorithm using two nonlinear operators which approximates
the photoreceptor. The two first layers of the retina, photoreceptor and OPL, have been modeled
and successfully used for face recognition under difficult lighting conditions \cite{vu2009retina}.
In \cite{benoit2005}, Benoit and Caplier modeled all the three layers for moving contours enhancement.
Our algorithm is inspired by the performance of the bipolar cells.

\subsection{Details of the proposed method}
In fact, there are two types of bipolar cells, called ON and OFF. The ON bipolar cells take into
account the difference of photoreceptor and horizontal cell responses, whereas the OFF bipolar
cells compute the difference of horizontal and photoreceptor cells. More precisely, if we apply
a DoG filter on an image for simulating the bipolar cells, a ``map'' with positive and negative
values will be obtained. Within this resultant ``map'', the positive values and the absolute of
the negatives values correspond respectively to the responses of the ON and OFF bipolar cells.
The proposed algorithm is inspired by this ``natural'' performance of the human visual system in
extracting image details, but also by the detailed analysis on the properties of the DoG filter itself.

The DoG filter is often used to approximate a LoG filter (Laplacian of Gaussian) due to its
low computational cost. It calculates the second spatial derivative of an image. In areas
where the image has a constant intensity, the filter response will be zero. Wherever an
intensity change occurs, the filter will give a positive response on the darker side and
a negative response on the lighter side. At a reasonably sharp edge between two regions of
uniform but different intensities, the filter response will be:
\begin{itemize}
  \item zero at a long distance from the edge
  \item positive on one side of the edge
  \item negative on the other side
  \item zero at some point in between, on the edge itself
\end{itemize}

\begin{figure}[htbp]
\centering
\includegraphics[width=0.2\columnwidth]{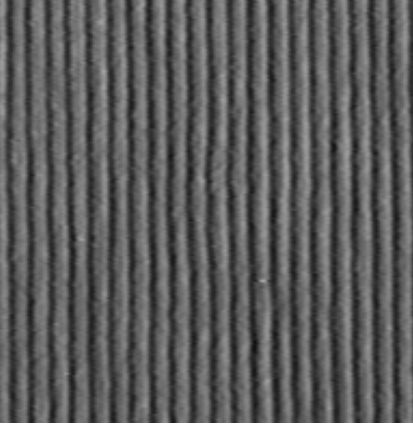}
\hspace{2mm}
\includegraphics[width=0.2\columnwidth]{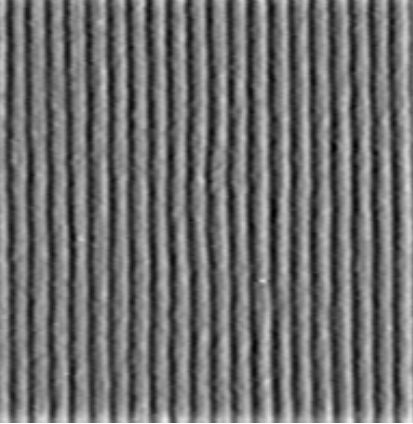}
\hspace{2mm}
\includegraphics[width=0.2\columnwidth]{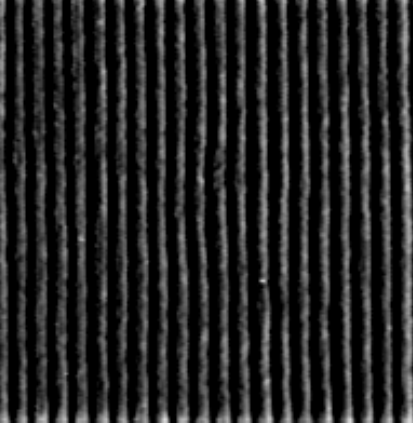}
\includegraphics[width=0.2\columnwidth]{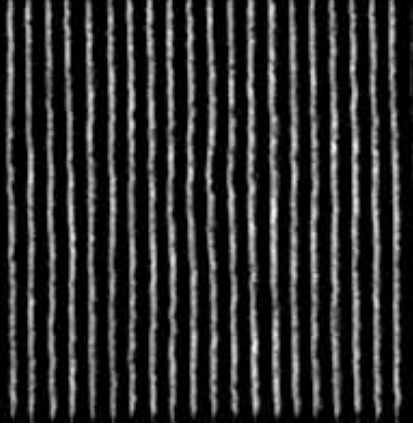}
\caption{Proposed processing chain: $I_{in}, I_{bf}, I^+_{bf}, I^-_{bf}$ (left to right).}
\label{fig:methex}
\end{figure}

Roughly speaking, the DoG filter can split the image details alongside two sides of the edge.
Keeping in mind all these properties of the DoG filter, we propose a novel two-step preprocessing
technique, as follows (Fig. \ref{fig:methex}):
        	
\textbf{Step 1:} given an image $I_{in}$, it is first filtered by a band-pass Difference of Gaussians
(DoG) filter which mimics the performance of bipolar cells:
\begin{eqnarray}
I_{bf}=DoG*I_{in}
\end{eqnarray}
where the $DoG$ kernel is given by:
\begin{eqnarray}
DoG=\frac{1}{2\pi\sigma_{1}^2} e^{-\frac{x^2+y^2}{2{\sigma}^2_{1}}}-\frac{1}{2\pi\sigma_{2}^2} e^{-\frac{x^2+y^2}{2{\sigma}^2_{2}}}
\end{eqnarray}
and where $\sigma_{1}$ and $\sigma_{2}$ correspond to the standard deviations of the
low pass filters modeling photoreceptors and horizontal cells.

\textbf{Step 2:} the responses at bipolar cells are then decomposed into two ``maps''
corresponding to the image details alongside the two sides of the image edge (these
decomposed ``maps'' also correspond to the responses on the ON and OFF cells):
\begin{eqnarray}
I^+_{bf}(p)=\left\{ \begin{array} {lll}
             I_{bf}(p) & if & I_{bf}(p) \ge \epsilon\\
			 0 & \multicolumn{2}{c}{otherwise}
			\end{array} \right.
\end{eqnarray}
\begin{eqnarray}
I^-_{bf}(p)=\left\{ \begin{array} {lll}
             |I_{bf}(p)| & if & I_{bf}(p) \le -\epsilon\\
			 0 & \multicolumn{2}{c}{otherwise}
			\end{array} \right.
\end{eqnarray}
where the term $bf$ refers to ``Biologically-inspired Filtering'', $p$ refers to a
considered pixel, $\epsilon$ is slightly larger than zero to provide some stability
in uniform regions: we do not take into account the uniform areas since these areas
often contain noise rather than useful texture information.

Then, in the feature extraction step, instead of the input image $I_{in}$, features are
first extracted from the two images, $I^-_{bf}$ and $I^+_{bf}$, and then combined together.
The resultant feature vector is considered as the descriptor of $I_{in}$. For example,
with the conventional LBP method, the two histograms of LBP codes estimated from $I^-_{bf}$
and $I^+_{bf}$ are concatenated and considered as the texture representation of $I_{in}$.

Our algorithm is different from Local Ternary Patterns (LTP) \cite{tan2007a} in several aspects.
First, LTP splits the LBP codes into their positive and negative parts by comparing directly the
pixel intensities, while we split the image filtered by DoG based on its edges. While LTP
encodes the first order pixel-wise information, our method computes the second spatial derivative.
Moreover, as a \emph{preprocessing} technique, our algorithm can be easily combined with
different \emph{feature extraction} approaches, including LTP, to get more efficient methods.
Also, we will show that the combination of the BF filter with the conventional LBP method considerably outperforms LTP.

\subsection{Properties}\label{ss:algo_pro}
The proposed preprocessing has the following advantages:
\begin{itemize}
\item Robust to illumination and noise: the band-pass DoG filter removes both high frequency noise and low frequency illumination.

\item Discriminative: with our technique, image details alongside two sides of the edges are obtained. We will experimentally show that features, which are extracted from these ``maps'' and combined together, convey richer information about object than features being extracted from an input image.

\item Rotation insensitivity: our filter is isotropic and discards all orientation information, and is therefore independent to rotation. 

\item Low computational time: the proposed algorithm is extremely fast. With un-optimized Matlab code running on a laptop of CPU Intel Core i5 1.7Ghz (2G Ram), it takes only less than $0.9s$ to process 1000 images of 128$\times$128 pixels or $1.9s$ to process 1000 images of 200$\times$200 pixels.
\end{itemize}

\section{Experimental Validation}\label{ss:result}
\subsection{Experimental Settings}
\subsubsection{Databases}
The effectiveness of the proposed method is assessed by a series of
experiments on three large and representative databases: Outex \cite {ojala2002outex}, CUReT (Columbia-Utrecht Reflection and Texture) \cite{curet1999}, and UIUC \cite{lazebnik2005pami}.

\begin{figure}[htbp]
\centering
\includegraphics[width=0.8\columnwidth]{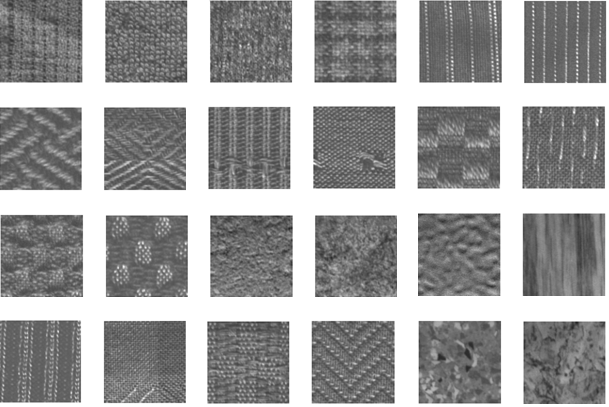}
\caption{Texture images with the illumination condition ``inca'' and zero degree rotation
angle from the 24 classes of textures on the Outex database.}
\label{fig:outex}
\end{figure}

The Outex database (examples are shown in Fig. \ref{fig:outex}) contains textural images captured
from a wide variety of real materials. We consider the two commonly used test suites, Outex\_TC\_00010 (TC10) and Outex\_TC\_00012 (TC12), containing 24 classes of textures which
were collected under three different illuminations (``horizon'', ``inca'',
and ``t184'') and nine different rotation angles (0$^{\circ}$, 5$^{\circ}$, 10$^{\circ}$, 15$^{\circ}$, 30$^{\circ}$, 45$^{\circ}$, 60$^{\circ}$, 75$^{\circ}$ and 90$^{\circ}$). 

\begin{figure}[htbp]
\centering
\includegraphics[width=0.6\columnwidth]{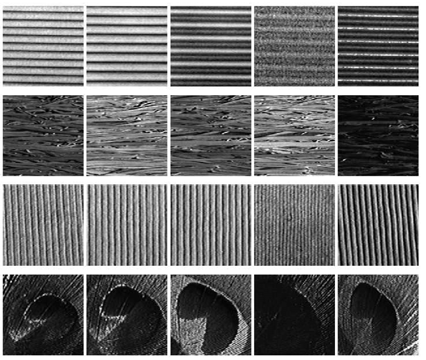}
\caption{Examples of texture images from the CUReT database.}
\label{fig:curet}
\end{figure}

The CUReT database (examples are shown in Fig. \ref{fig:curet}) contains 61 texture classes, each
having 205 images acquired at different viewpoints and
illumination orientations. There are 118 images shot from a
viewing angle of less than 60$^{\circ}$. From these 118 images, as in
\cite{varma2009pami,clbp2010tip}, we selected 92 images, from which a sufficiently large region
could be cropped (200$\times$200) across all texture classes.
All the cropped regions are converted to grey scale.

\begin{figure}[htbp]
\centering
\includegraphics[width=0.6\columnwidth]{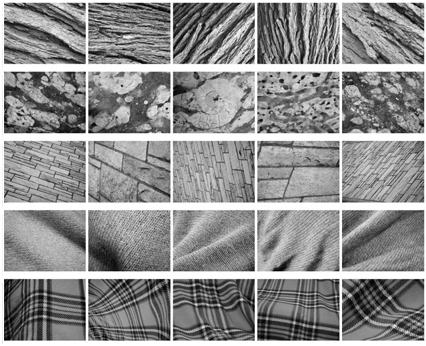}
\caption{Examples of texture images on the UIUC database.}
\label{fig:uiuc}
\end{figure}

The UIUC texture database includes 25 classes with 40 images
in each class. The resolution of each image is 640$\times$480. The database contains materials imaged under significant viewpoint variations (examples are shown in Fig. \ref{fig:uiuc}).

\subsubsection{Feature Extraction}
To validate the proposed preprocessing algorithm, we evaluate how far it can improve the performance of texture classification systems based on several popular features, including SIFT, LBP-based descriptors, WLD (Weber Local Descriptor) \cite{chen2010weber}, and LBC (Local Binary Count - a recent LBP variant) \cite{clbc2012tip}. Note that, the combination of BF with other more powerful features such as the extended LBP method \cite{liu2012imavis} is beyond the scope of this paper and will be considered in future work.

Concerning LBP, we use the completed model proposed in \cite{clbp2010tip}, which gathers three individual operators: CLBP-Sign (CLBP\_S) which is equivalent to the conventional LBP, CLBP-Magnitude (CLBP\_M) which measures the local variance of magnitude, and CLBP-Center (CLBP\_C) which extracts the local central information. Similarly, the completed LBC model with three different individual operators is considered. When using LBP and LBC features, only rotation invariant uniform patterns are considered \footnote{The rotation invariant uniform patterns are often denoted with the subscript $^{riu_2}$, e.g., LBP$^{riu_2}$ or LBC$^{riu_2}$, but for simplicity in this paper, they are simply denoted by their names, without $^{riu_2}$, e.g., LBP or LBC.}.

We also apply the combination strategies presented in \cite{clbp2010tip}. There are two ways to combine the CLBP\_S and CLBP\_M codes: by concatenation or jointly. In the first way, the histograms of the CLBP\_S and CLBP\_M codes are computed separately, and then concatenated together. This CLBP scheme is referred to as ``CLBP\_S\_M''. In the second way, a joint 2D histogram of the CLBP\_S and CLBP\_M codes is calculated and denoted as ``CLBP\_S/M''.

The three operators, CLBP\_S, CLBP\_M and CLBP\_C (resp. CLBC\_S, CLBC\_M and CLBC\_C in the completed LBC model), can be combined in two ways, jointly or hybridly. In the first way, a 3D joint histogram of them is built and denoted by ``CLBP\_S/M/C''. In the second way, a 2D joint histogram, ``CLBP\_S/C'' or ``CLBP\_M/C'' is built first, and then is converted to a 1D histogram, which is then concatenated with CLBP\_M or CLBP\_S to generate a joint histogram, denoted by ``CLBP\_M\_S/C'' or ``CLBP\_S\_M/C''.

The WLD method \cite{chen2010weber} consists of two components: differential excitation and orientation. The differential excitation component is a function of the ratio between two terms: one is the relative intensity differences of a current pixel against its neighbors; the other is the intensity of the current pixel. The orientation component is the gradient orientation of the current pixel. For a given image, the two components are used to construct a concatenated WLD histogram.

We therefore apply these schemes on the original texture images and on the preprocessed images, obtained as presented in Section \ref{ss:algo}.

When using SIFT (in this paper, we used the original SIFT introduced by Lowe \cite{lowe2004}), given an image, co-variant regions (patches around keypoints) are first detected and then processed by the BF filter to generate richer information. Finally, SIFT descriptors are computed from ``generated patches'' and used to represent corresponding co-variant regions in the original image. In other words, feature keypoints are detected in original image while feature descriptors are computed in images filtered by the BF approach.

\subsubsection{Similarity Measure and Classifier}
Classification rates are reported using the simple nearest neighbor classifier. 

For all CLBP, CLBC and WLD descriptors, the $\chi^2$ distance is used to measure the similarity between two texture images. If $H = \{h_i\}$ and $K = \{k_i\}, (i = 1, 2,...,n)$ denote two histograms corresponding to the representations of two images, the $\chi^2$ distance is calculated as: $\chi^2 (H,K) = \sum_{i=1}^n \frac{(h_i-k_i)^2}{h_i+k_i}$.

Concerning SIFT, keypoints from two images are compared and matched using the L2 norm of the difference between their descriptors as measure. The final similarity score between two considered images is the average value of distance between their matched keypoints (when there are any matched keypoints between two images, the score is set very high).

\subsection{Parameter Exploration}
In this section, we study how the parameters of the BF filter influence final performance. Parameters to be chosen include the two standard deviations $\sigma_{1}$ and $\sigma_2$ defining the low and high cutoff frequencies of the band pass $DoG$ filter, and the threshold $\epsilon$. A critical constraint is $\sigma_{1} < \sigma_2$ (recall that these correspond respectively to the parameters of the blurring filters being performed in photoreceptors and horizontal cells).

The experiments described in this section were conducted on the Outex database. We compute the average classification rates on the three test suites (TC10, TC12t and TC12h) with different parameters of the BF filter. For each texture descriptor or combination strategy in the completed LBP/LBC model used (each descriptor was also tested with different parameters itself), 150 BF filters with different parameters were evaluated:
\begin{itemize}
  \item $\sigma_{1}\in\{0.5,0.75,1,1.25,1.5\}$
  \item $\sigma_{2}\in\{2,3,4,5,6\}$
  \item $\epsilon \in\{0.05,0.1,0.15,0.2,0.25,0.3\}$
\end{itemize}

\begin{figure*}[htbp]
\centering
\subfigure[CLBC\_S$_{R=2}$]
{
\includegraphics[width=0.23\textwidth]{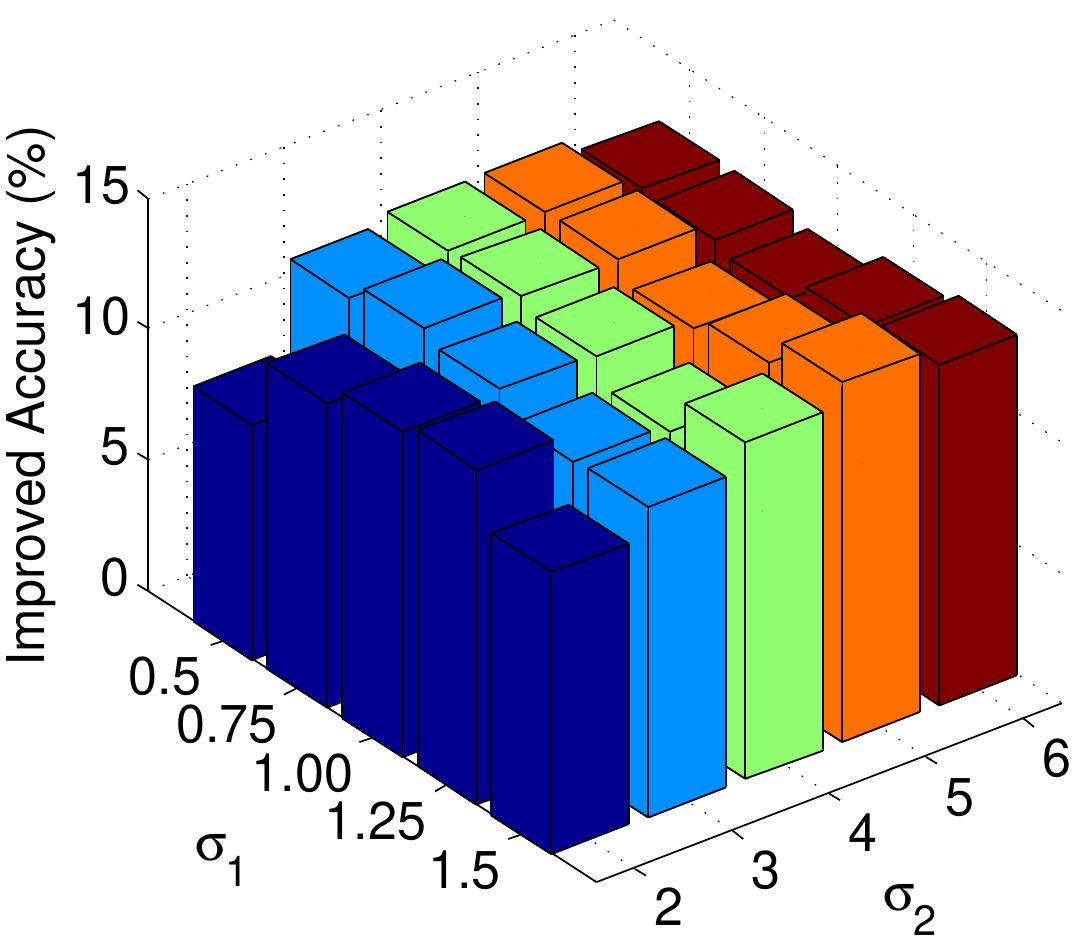}
}
\subfigure[CLBC\_M$_{R=2}$]
{
\includegraphics[width=0.23\textwidth]{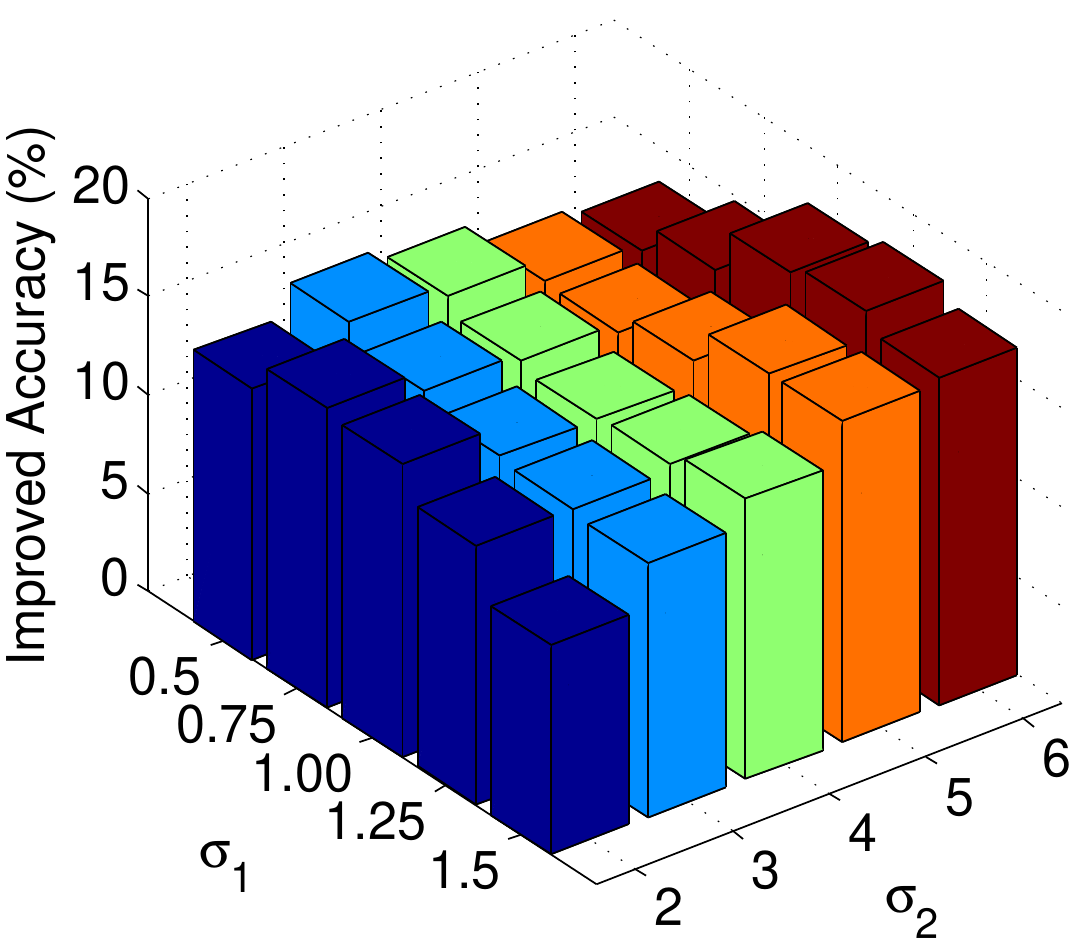}
}
\subfigure[CLBC\_S/M$_{R=2}$]
{
\includegraphics[width=0.23\textwidth]{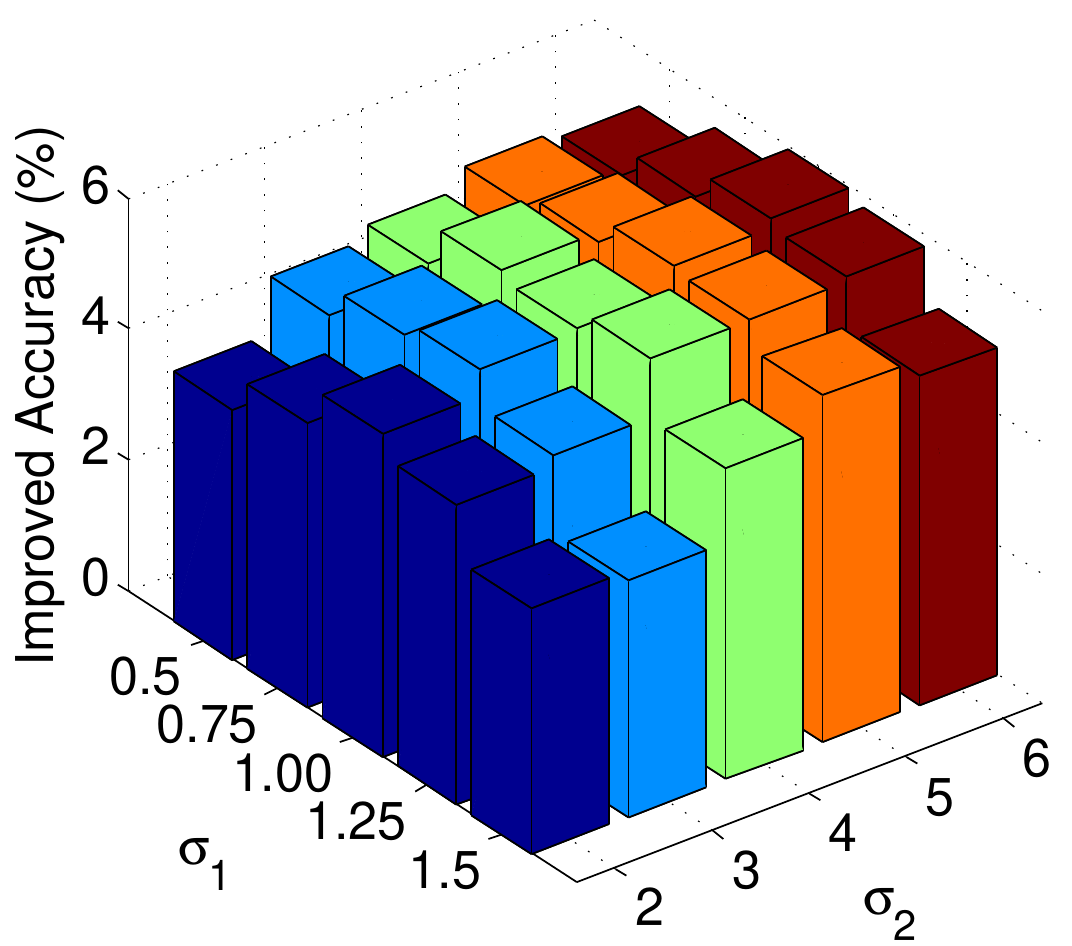}
}
\subfigure[CLBC\_S/M/C$_{R=2}$]
{
\includegraphics[width=0.23\textwidth]{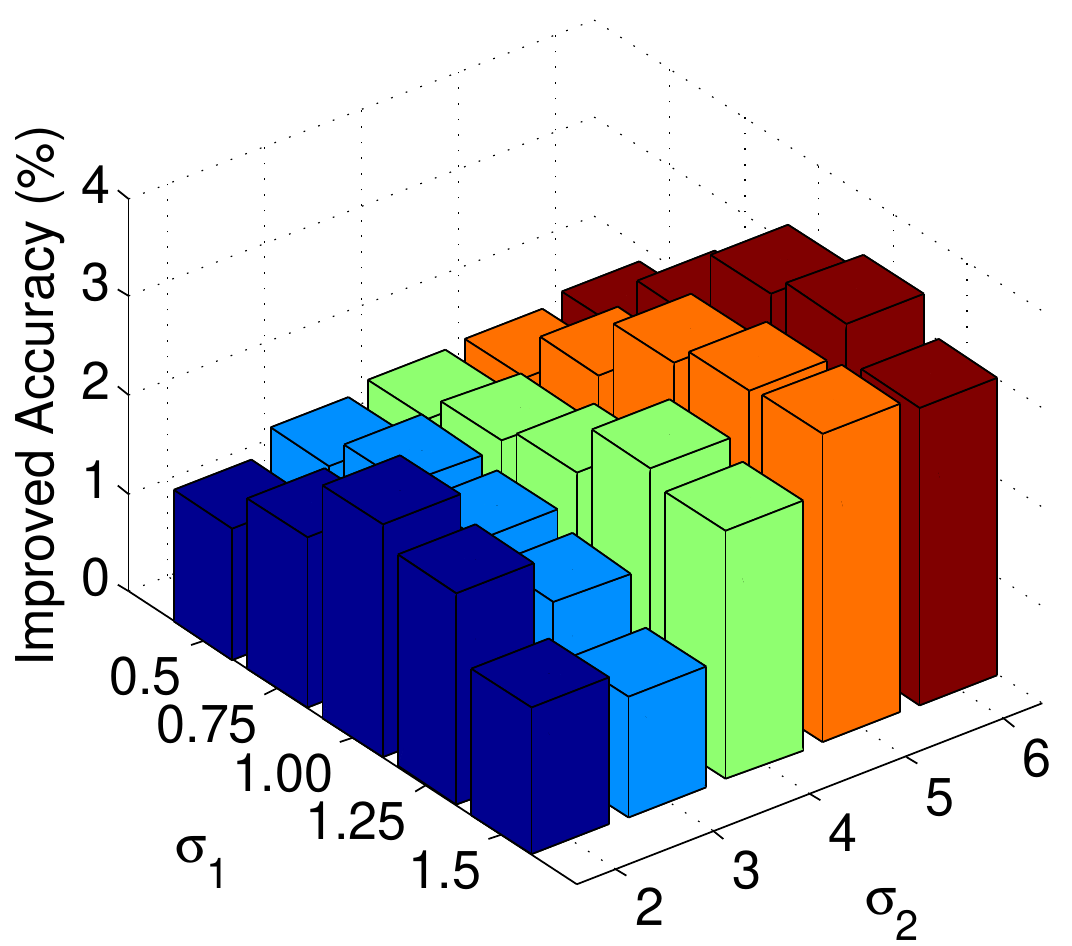}
}\\

\subfigure[CLBP\_S/C$_{R=1}$]
{
\includegraphics[width=0.23\textwidth]{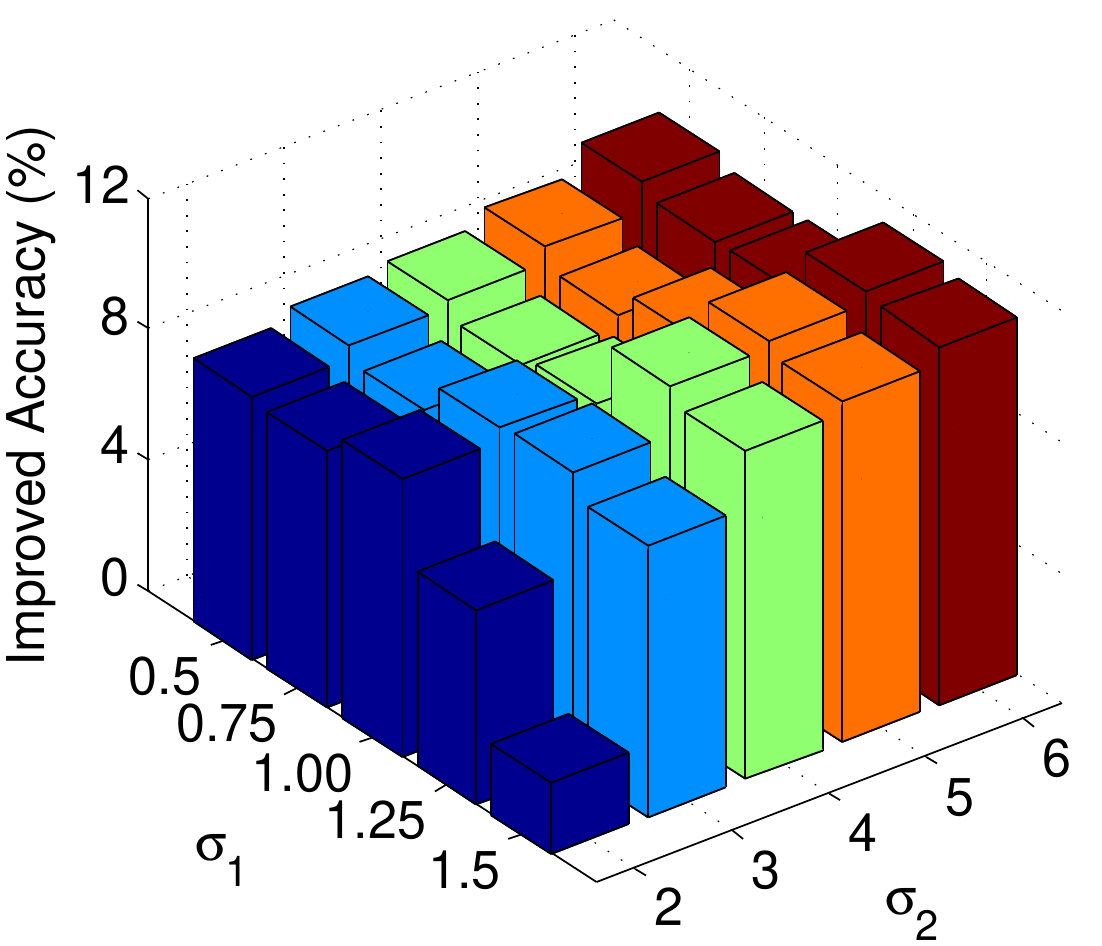}
}
\subfigure[CLBP\_M/C$_{R=2}$]
{
\includegraphics[width=0.23\textwidth]{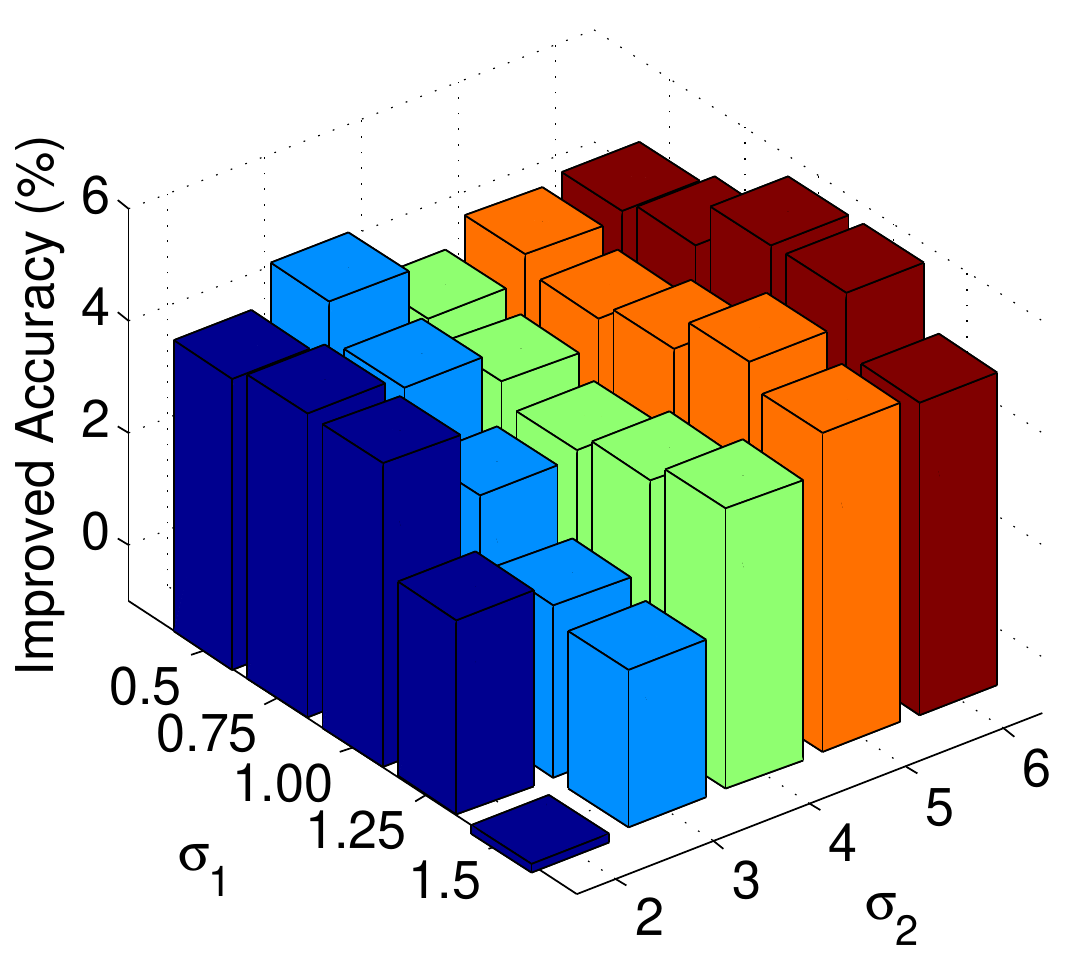}
}
\subfigure[CLBP\_S\_M/C$_{R=2}$]
{
\includegraphics[width=0.23\textwidth]{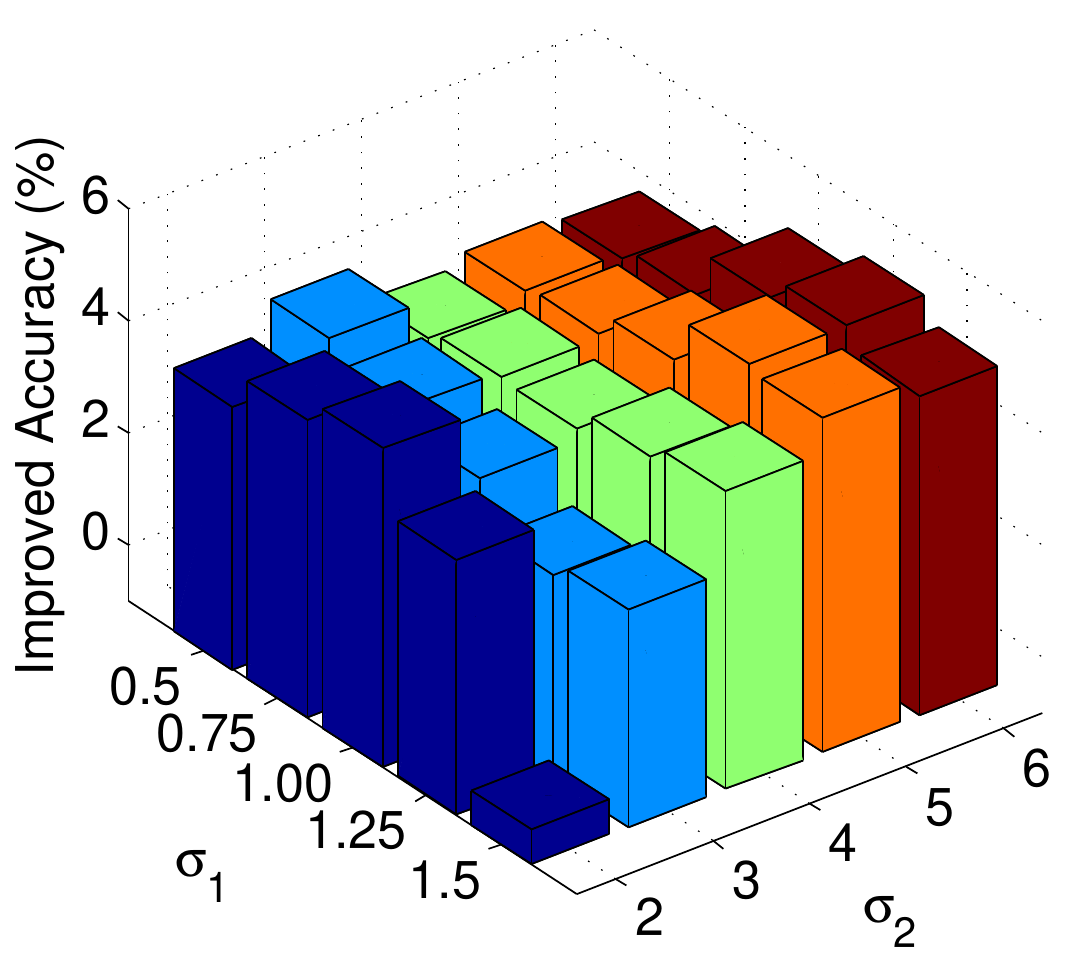}
}
\subfigure[WLD]
{
\includegraphics[width=0.23\textwidth]{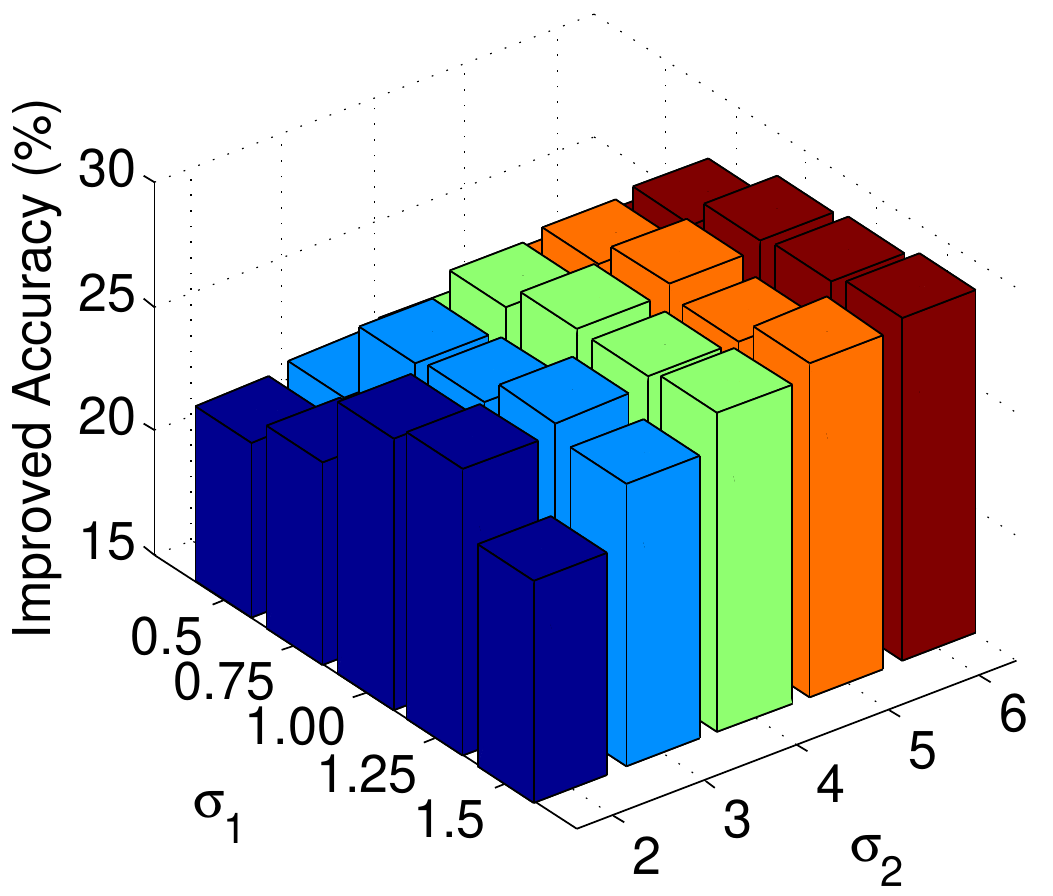}
}
\caption{Improved classification accuracy obtained with different standard deviations ($\sigma_1$ and $\sigma_2$) of the BF filter. These rates are calculated by averaging classification rates with different $\epsilon$ on the three test suites (TC\_10, TC\_12 ``t'', TC\_12 ``h'') of the Outex database.}
\label{fig:para1}
\end{figure*}

\subsubsection{Determining $\sigma_1$ and $\sigma_2$}

For each pair of $\sigma_1$ and $\sigma_2$, the average classification rates are calculated for varying $\epsilon$. Fig. ~\ref{fig:para1} shows the \emph{improved} classification rates obtained when the proposed BF filter is used with different descriptors and combination strategies (due to space limitations, the most informative results are depicted). In a few cases when the values of $\sigma_1$ and $\sigma_2$ are very close (see Fig. ~\ref{fig:para1} (f) and (g) with $\sigma_1=1.5, \sigma_2=2,$), the BF filtering leads to slight improvements. This can be explained by the fact that when $\sigma_1$ is slightly smaller than $\sigma_2$, a high amount of image information is ignored. More precisely, in such cases, the information at numerous frequencies (low and high) is removed. In all the other cases, our BF filters always result an in important gain of classification accuracy for all evaluated descriptors (further details will be discussed in Section IV.D). When using CLBC\_S/M/C, CLBP\_S/M/C or WLD (Fig. ~\ref{fig:para1} (d) and (h)), the performance of the BF filter with $\sigma_1=0.5$ is slightly worse than the others. Therefore, in general, we propose to use $\sigma_1\in\{0.7,1.3\}$ and $\sigma_2\in\{3,6\}$. Indeed, in our tests, we obtained the best results with $\sigma_1=1.25$ and $\sigma_2=5,6$ (refer to Section IV.D).

\subsubsection{Determining $\epsilon$}

\begin{figure}[htbp]
\centering
\includegraphics[width=0.75\columnwidth]{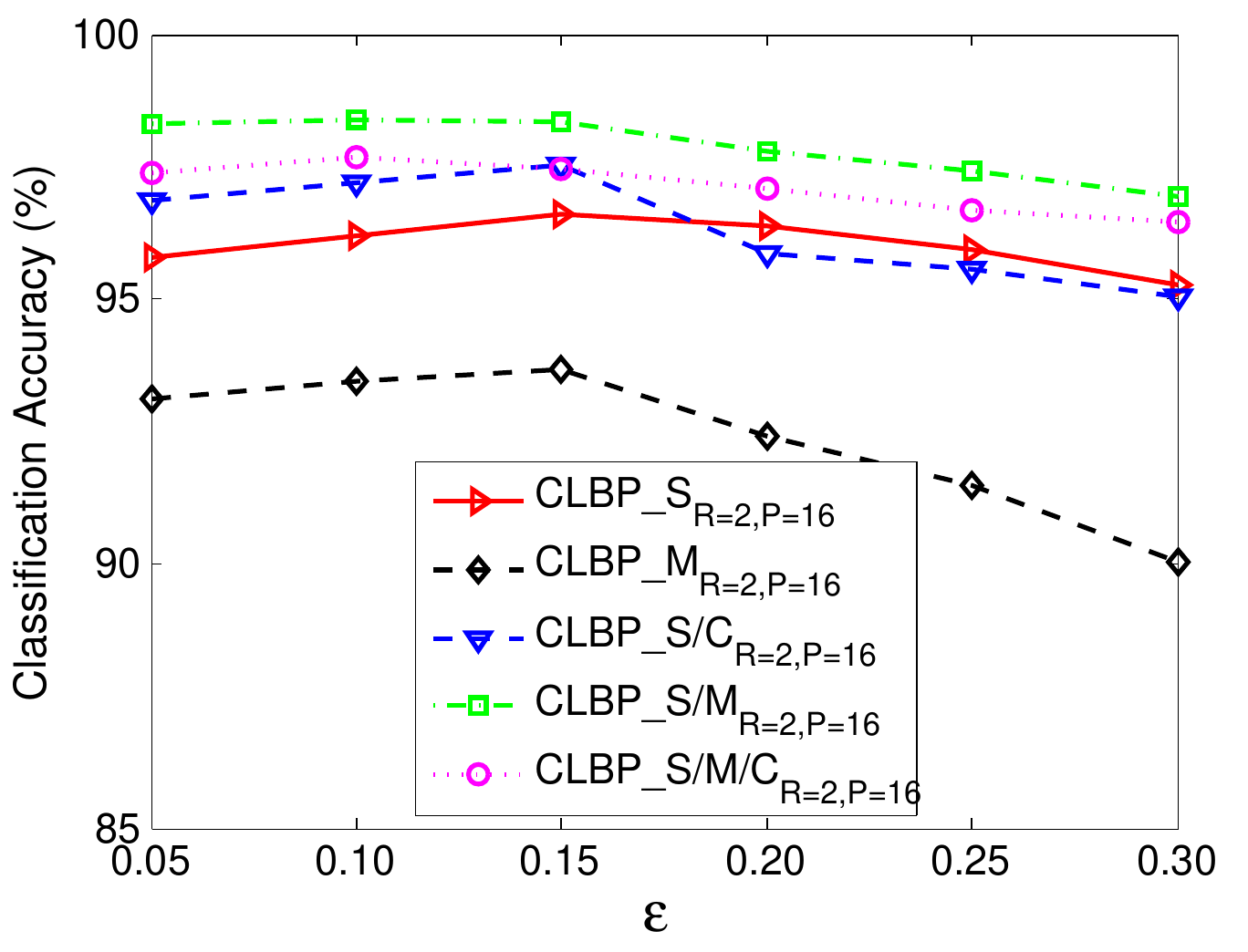}\\
\vspace{+3mm}
\includegraphics[width=0.75\columnwidth]{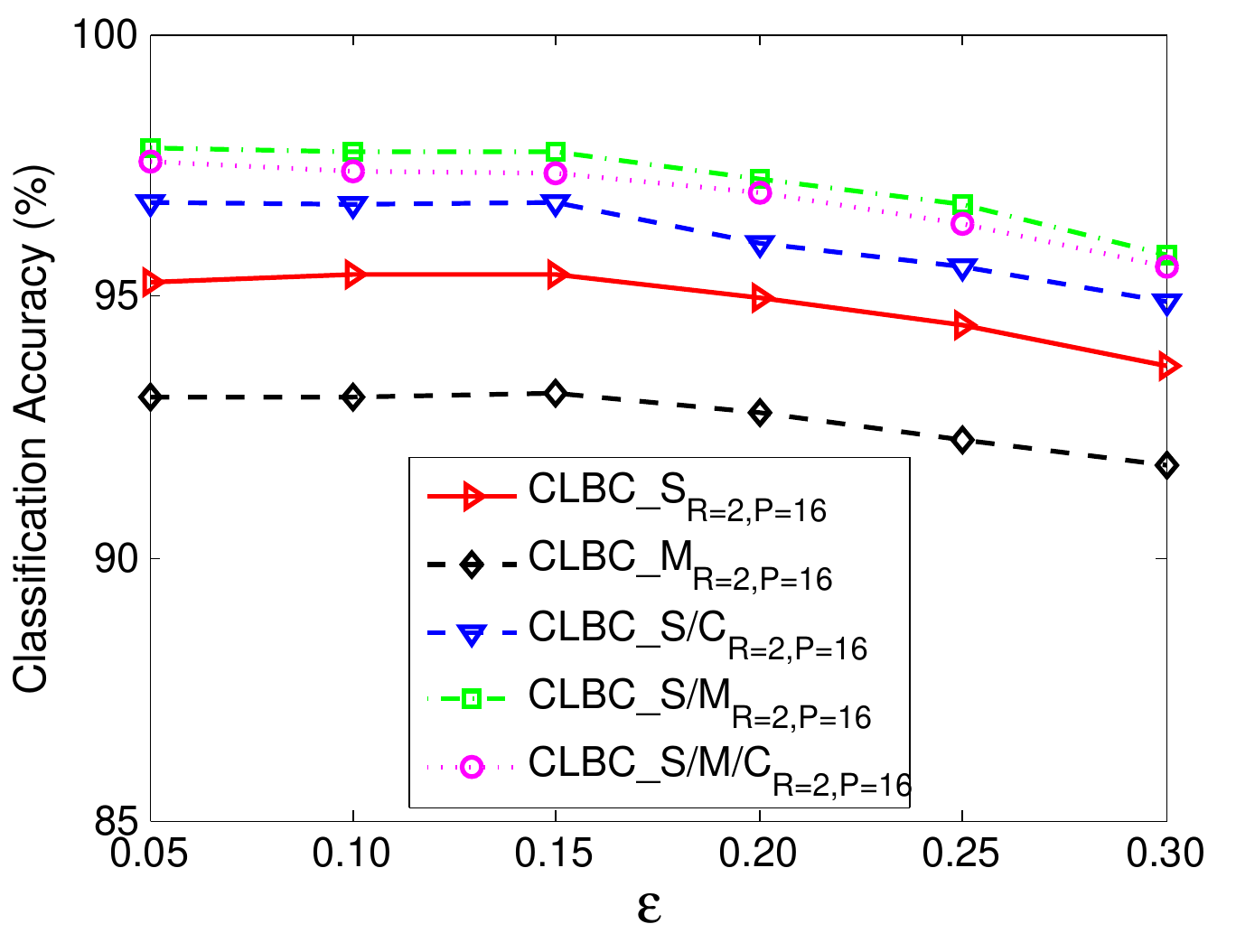}
\caption{Classification rates for different descriptors with different values of the threshold $\epsilon$. These rates are calculated by averaging classification rates on the three test suites (TC\_10, TC\_12 ``t'', TC\_12 ``h'') of the Outex database with different values of $\sigma_1$ and $\sigma_2$.}
\label{fig:para2}
\end{figure}

Recall that the threshold $\epsilon$ being slightly larger than zero is used to provide some stability
in uniform regions (the uniform areas containing noise rather than useful texture information are not taken into account). To determine $\epsilon$, we compute the average of classification rates across different $\epsilon$: for each value of $\epsilon$, we use 12 pairs of ($\sigma_1,\sigma_2$), $\sigma_1=0.75,1.00,1.25$ and $\sigma_2=3,4,5,6$. As can be seen from Fig. \ref{fig:para2}, the performance of the BF filter with $\epsilon = 0.05, 0.1, 0.15$ is similar (the BF filter with $\epsilon=0.15$ performs slightly better the others) while it begins to drop when $\epsilon \geq 0.2$. This can be explained by the fact that when the value of $\epsilon$ is too high, useful information is lost.

In conclusion, the BF filtering performs well with different parameters: $\sigma_1\in [0.7,1.3]$ and $\sigma_2\in[3,6]$, $\epsilon \in[0.05, 0.15]$. We obtained the best results with $\sigma_1=1.25$, $\sigma_2=5,6$, and $\epsilon = 0.15$ (when CLBP and CLBC are used). However, with the main goal of showing that our preprocessing technique is very efficient in general, in the rest of this paper, without exception, we will report the results with the default parameters: $\sigma_1=1$, $\sigma_2=4$, and $\epsilon = 0.1$.

\subsection{Comparison with other Preprocessing Algorithms}
In order to show the advantage of our algorithm, this section compares its performance with other preprocessing techniques. It would be worth noting that in the literature, to improve the overall quality of texture classification, researchers often focus on improving one of (or both) the two steps of representation and classification, and to the best of our knowledge, there does not exist any work considering properly preprocessing techniques. We propose to compare here the BF filter with several existing preprocessing techniques: (1) ``traditional'' gamma correction method (it is shown in \cite{liao2009tip} that histogram equalization (HE) degrades in general the performance of texture descriptor, thus we do not compare HE and BF here); (2) the typical DoG filter without dividing the filtered image into two parts as our BF method; and (3) Gaussian derivative filters (zeroth-order, first order and second order derivative filters). For all sorts of considered filters, we compute the classification rates with different parameters (e.g., different scales for Gaussian filters) and use their best classification rates to compare with our ``mean'' results (we do not use the optimal parameters).

We also decomposed the image filtered by Gaussian first derivative filter (gradient magnitude map) regarding the pixel gradient orientation and then extract the features on those decomposed images. However, contrary to face recognition \cite{vu2012pr,vu2013tifs}, this idea does not work for texture classification. This is due to the fact that considered texture images contain rotation transforms and dividing them across ``absolute'' orientations leads to sensitivity to rotation. More precisely, Gaussian first derivative filter is defined as:
\begin{equation}
    DtG(x,y,\sigma)=\sqrt{(G_x(x,y,\sigma))^2+(G_y(x,y,\sigma))}
\end{equation}
where $ G_x(x,y,\sigma)=\frac{dG(x,y,\sigma)}{dx}=-\frac{x}{2\sigma^3\pi}e^{-\frac{x^2+y^2}{2\sigma^2}}$ and
$G_y(x,y,\sigma)=\frac{dG(x,y,\sigma)}{dy}=-\frac{y}{2\sigma^3\pi}e^{-\frac{x^2+y^2}{2\sigma^2}}$.

\begin{figure}[htbp]
\centering
\subfigure[Outex\_TC\_00010]
{
\includegraphics[width=0.95\columnwidth]{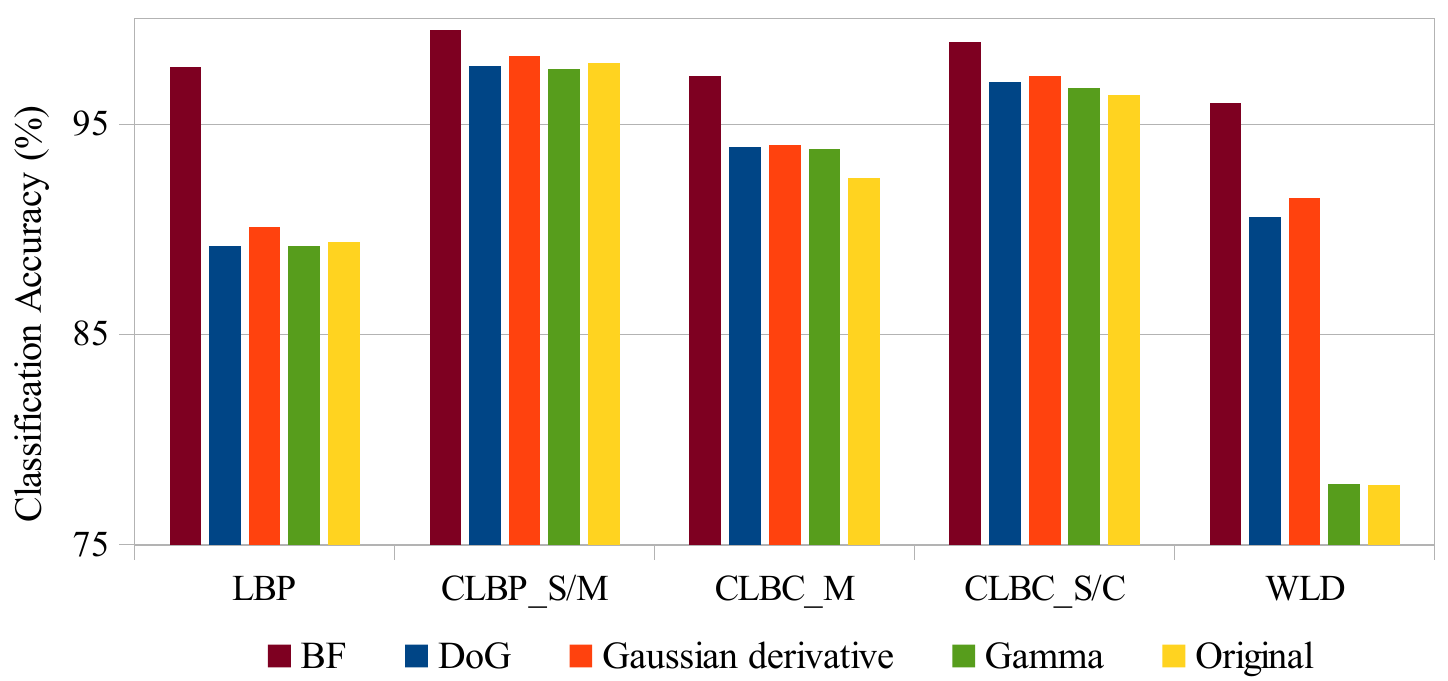}
}
\subfigure[Outex\_TC\_00012 (t184)]
{
\includegraphics[width=0.95\columnwidth]{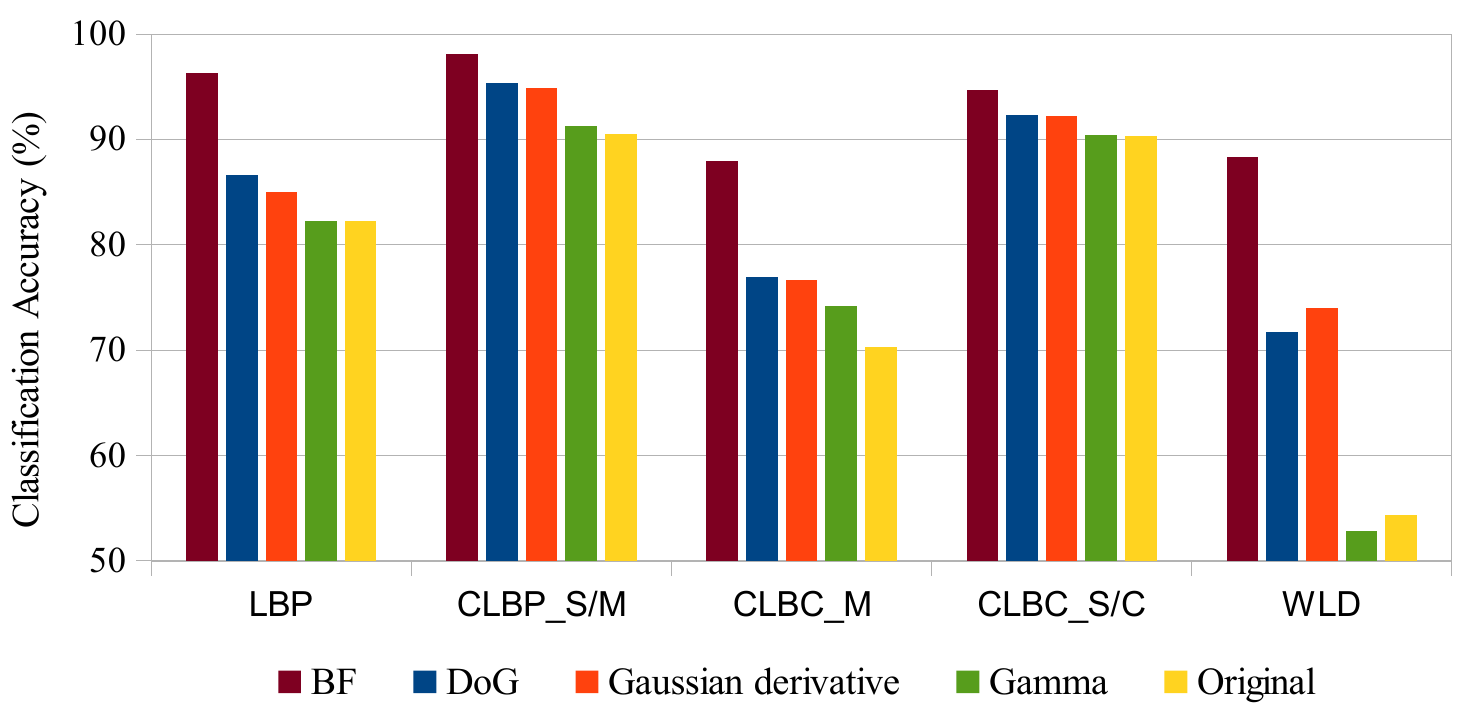}
}
\subfigure[Outex\_TC\_00012 (horizon)]
{
\includegraphics[width=0.95\columnwidth]{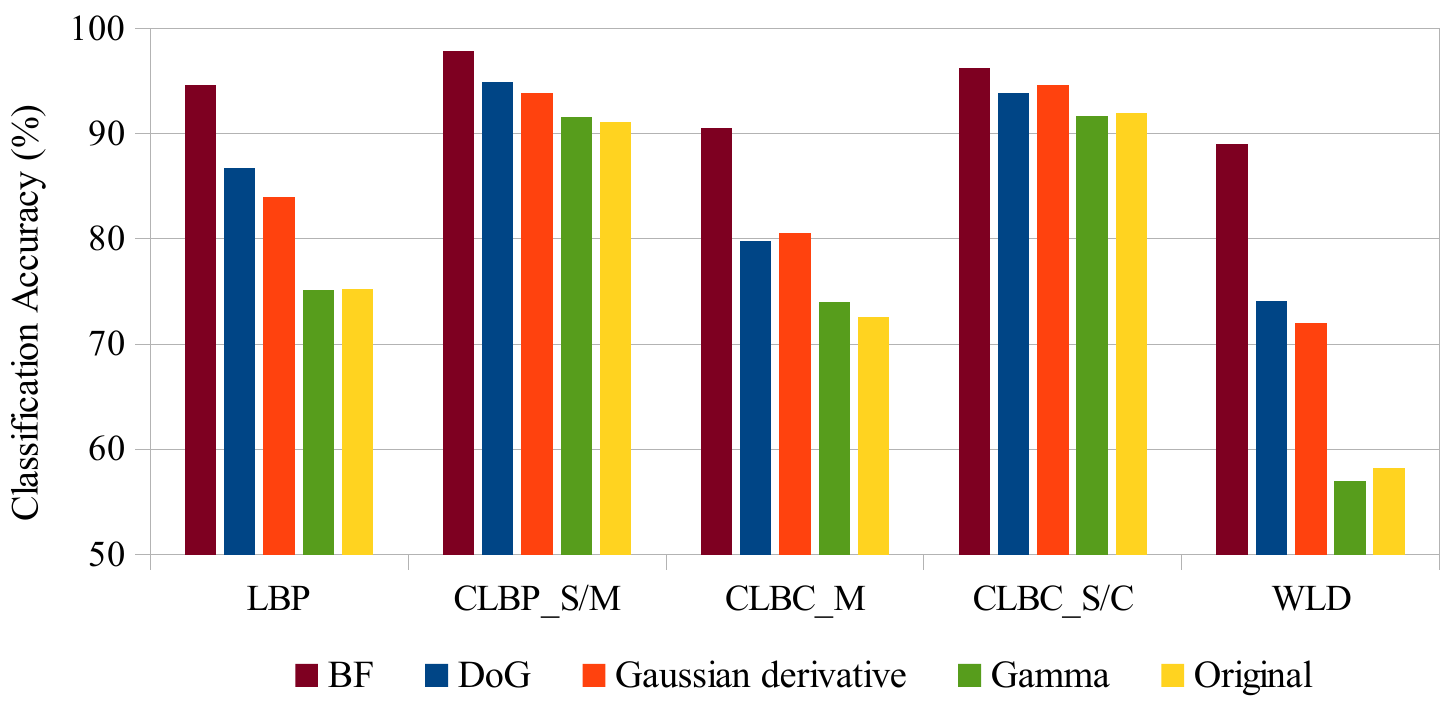}
}
\caption{Comparison of different preprocessing techniques.}
\label{fig:difpp}
\end{figure}

Using Gabor filters as preprocessing (the global mean of the responses of Gabor filters) was also considered but in our experiments, they perform worse than Gaussian derivative filters. The classification results of different preprocessing techniques are shown in Fig. \ref{fig:difpp} (only the most informative results are depicted). As can be seen, preprocessing techniques can enhance the performance of texture classification, and in all cases, our BF filter always performs the best, showing the advantage of the proposed preprocessing approach.

\subsection{Results on the Outex Database}
This section presents and discusses in more detail the results obtained on the Outex database. We first show how far our preprocessing algorithm can improve the texture descriptors ``in general'': the default parameters ($\sigma_1=1$, $\sigma_2=4$, and $\epsilon = 0.1$) are used. Then, with the optimal parameters, we show that combining our preprocessing approach with known texture descriptors can outperform various state-of-the-art texture classification systems. The results obtained when combining BF with CLBP will be analyzed in detail while other results will de discussed more quickly.

\subsubsection{Improving the performance of the completed LBP model}
For each method considered, three classification rates are computed with different parameters: $P$ being the total number of involved neighbors per pixel and $R$ the radius of the neighborhood.

\begin{table*}[htbp]

\begin{center}
\caption{CLASSIFICATION RATES OBTAINED ON THE OUTEX DATABASE WHEN USING CLBP AS DESCRIPTOR.}
\begin{tabular}{l|rrrr|rrrr|rrrr}
\hline \hline
&\multicolumn{4}{c}{$R=1, P=8$}&\multicolumn{4}{c}{$R=2, P=16$}&\multicolumn{4}{c}{$R=3, P=24$}\\
\hline
& \multicolumn{1}{c}{TC10} & \multicolumn{2}{c}{TC12} & {Average} & \multicolumn{1}{c}{TC10} & \multicolumn{2}{c}{TC12} & {Average} & \multicolumn{1}{c}{TC10} & \multicolumn{2}{c}{TC12} & \multicolumn{1}{c}{Average} \\
& & \multicolumn{1}{c}{t184} & \multicolumn{1}{c}{horizon} & & & \multicolumn{1}{c}{t184} & \multicolumn{1}{c}{horizon} & & & \multicolumn{1}{c}{t184} & \multicolumn{1}{c}{horizon} & \\
\hline
\hline
LTP & 94.14 & 75.88 & 73.96 & 81.33 & 96.95 & 90.16 & 86.94 & 91.35 & 98.20 & 93.59 & 89.42 & 93.74 \\
\hline
\hline
CLBP\_S      & 84.81 & 65.46 & 63.68 & 71.31 & 89.40 & 82.26 & 75.20 & 82.28 & 95.07 & 85.04 & 80.78 & 86.96\\
BF + CLBP\_S & 96.17 & 92.92 & 92.75 & 93.95 & 97.73 & 96.29 & 94.58 & 96.20 & 99.11 & 96.62 & 93.15 & 96.29\\
\textit{Gain}& \textit{+11.36} & \textit{+27.46} & \textit{+29.07} & \textbf{\textit{+22.64}} & \textit{+8.33} & \textit{+14.03} & \textit{+19.38} & \textbf{\textit{+13.92}} & \textit{+4.04} & \textit{+11.58} & \textit{+12.37} & \textbf{\textit{+9.33}}\\
\hline
\hline
CLBP\_M      & 81.74 & 59.30 & 62.77 & 67.93 & 93.67 & 73.79 & 72.40 & 79.95 & 95.52 & 81.18 & 78.65 & 85.11\\
BF + CLBP\_M & 91.43 & 79.40 & 83.59 & 84.81 & 97.29 & 89.47 & 93.59 & 93.45 & 97.08 & 92.78 & 94.28 & 94.71\\
\textit{Gain}& \textit{+9.69} & \textit{+20.10} & \textit{+20.82} & \textit{+16.08} & \textit{+3.62} & \textit{+15.68} & \textit{+21.19} & \textit{+13.50} & \textit{+1.56} & \textit{+11.60} & \textit{+15.63} & \textit{+9.60}\\
\hline
\hline
CLBP\_M/C      & 90.36 & 72.38 & 76.66 & 79.80 & 97.44 & 86.94 & 90.97 & 91.78 & 98.02 & 90.74 & 90.69 & 93.15\\
BF + CLBP\_M/C & 94.43 & 83.84 & 86.53 & 88.27 & 99.24 & 91.34 & 94.65 & 95.08 & 98.93 & 95.53 & 96.46 & 96.97\\
\textit{Gain}  & \textit{+4.07} & \textit{+11.46} & \textit{+9.87} &  \textit{+8.47} & \textit{+1.80} & \textit{+4.40} & \textit{+3.68} & \textit{+3.30} & \textit{+0.91} &  \textit{+4.79} &  \textit{+5.77} &  \textit{+3.82}\\

\hline
\hline
CLBP\_S\_M/C& 94.53 & 81.87 & 82.52 & 86.30 & 98.02 & 90.99 & 91.08 & 93.36 & 98.33 & 94.05 & 92.40 & 94.92 \\
BF + CLBP\_S\_M/C & 96.69 & 90.76 & 91.48 & 92.98 & 99.45 & 95.56 & 96.16 & 97.06 & 99.32 & 96.83 & 96.50 & 97.55 \\
\textit{Gain} & \textit{+2.16} & \textit{+8.89} & \textit{+8.96} & \textit{+6.68} & \textit{+1.43} & \textit{+4.57} &  \textit{+5.08} & \textit{+3.70} & \textit{+0.99} & \textit{+2.78} & \textit{+4.10} & \textit{+2.63}\\
\hline
\hline
CLBP\_S/M & 94.66 & 82.75 & 83.14 & 86.85 & 97.89 & 90.55 & 91.11 & 93.18 & 99.32 & 93.58 & 93.35 & 95.41\\
BF + CLBP\_S/M & 97.06 & 94.84 & 93.49 & 95.13 & 99.35 & \textbf{98.06} & \textbf{97.80} & \textbf{98.40} & \textbf{99.61} & \textbf{97.62} & 97.57 & 98.26\\
\textit{Gain} & \textit{+2.40} & \textit{+12.09} & \textit{+10.35} & \textit{+8.28} & \textit{+1.46} & \textit{+7.51} & \textit{+6.69} & \textit{+5.22} & \textit{+0.29} & \textit{+4.04} & \textit{+4.22} &\textit{+2.85}\\
\hline
\hline
CLBP\_S/M/C & 96.56 & 90.30 & 92.29 & 93.05 & 98.72 & 93.54 & 93.91 & 95.39 & 98.93 & 95.32 & 94.53 & 96.26\\
BF + CLBP\_S/M/C & \textbf{97.40} &  \textbf{95.28} & \textbf{94.58} & \textbf{95.51} & \textbf{99.50} & 96.87 & 96.62 & 97.67 & \textbf{99.63} & 97.52 & \textbf{97.71} & \textbf{98.29}\\
\textit{Gain} & \textit{+0.84} & \textit{+4.98} & \textit{+2.29} & \textit{+2.46} & \textit{+0.78} & \textit{+3.33} & \textit{+2.71} & \textit{+2.28} & \textit{+0.70} & \textit{+2.20} & \textit{+3.18} & \textit{+2.03}\\
\hline
\hline

\end{tabular}
\end{center}
\label{tab:outex1}
\end{table*}

Table I reports the experimental results of different methods,
from which we can get some interesting findings:

\begin{itemize}
\item For all considered features or combination strategies with their different parameter settings, using our algorithm as preprocessing results in important gains in performance. For example, when combining our BF filtering with the conventional LBP (CLBP\_S in Table I), with the three parameter configurations, the average improvements are respectively 22.64$\%$, 13.92$\%$, and 9.33$\%$. Similarly, when combining the BF filtering with CLBP\_M, the average improvements with the three parameters are respectively 16.08$\%$, 13.50$\%$, and 9.60$\%$.

\item With the same parameters, the \emph{simple} combination ``BF $+$ LBP'' \emph{outperforms all}  ``original'' combination schemes which must gather the information of sign, magnitude, and/or center (by term ``original'', we refer to algorithms which do not use our filtering as preprocessing). For example, with ($R=1, P=8$), the classification rate of ``BF $+$ LBP'' is 93.95\% while the results of CLBP\_S/M and CLBP\_S/M/C are respectively 86.85\% and 93.05\%. Similarly, with ($R=2, P=16$), the classification rate of ``BF $+$ LBP'' is 96.20\% while the results of CLBP\_S/M and CLBP\_S/M/C are respectively 93.18\% and 95.39\%.
\item `BF $+$ LBP'' clearly outperforms LTP with all considered parameters on all test suites.
\item To compare the complexities of these algorithms, three factors must be considered: preprocessing time, feature extraction time and feature size which affects the classification time. As can be seen in Table II, our preprocessing step requires only $0.87ms$ (resp. $1.91ms$) to process an image of 128$\times$128 (resp. 200$\times$200) pixels, this additional time being really small. When using the BF filter, the feature extraction time and feature size are doubled, but there are important gains in classification rates. The feature extraction time of ``BF $+$ LBP'' is similar to that of CLBP\_S\_M/C and CLBP\_S/M (see Table II). However, while the feature sizes of CLBP\_S\_M/C are 30, 54, and 78, the feature sizes of CLBP\_S/M are 100, 324, and 676 for ($R=1,P=8$), ($R=2,P=16$), and ($R=3,P=24$), respectively, the feature sizes of ``BF $+$ LBP'' are much smaller: they are 20, 36, and 52 for ($R=1,P=8$), ($R=2,P=16$), and ($R=3,P=24$) respectively.

\item ``BF + CLBP\_S'' outperforms the best ``original'' scheme, the CLBP\_S/M/C method. For instance, with (R=2, P=16), the classification rates of ``BF + CLBP\_S'' and CLBP\_S/M/C are 96.20\% and 95.39\% respectively. However, the feature dimensions of ``BF + CLBP\_S'' are considerably smaller than those of CLBP\_S/M/C. With the three parameter settings, the feature dimensions of ``BF + CLBP\_S'' are respectively 20, 36 and 52, while those of CLBP\_S/M/C are respectively 200, 648 and 1352. Thus, our classification step is faster, notably when texture classification is performed on large databases. In such systems, once a vector feature is extracted from the test image, it is compared and matched against (up to) thousands of other feature vectors corresponding to training samples in the database.

\end{itemize}

\begin{table}[]
\begin{center}
\caption{COMPARISON OF THE COMPLEXITY OF DIFFERENT TEXTURE REPRESENTATIONS.}
\begin{tabular}{l|rrrr}
\hline \hline
Method & FET$_1$ & FET$_2$ & Feat. Size & MT\\
\hline
\hline
BF & 0.87  & 1.91 & - & -\\
LBP$_{P=16}$ &      4.33 & 12.21 & 18 & 0.79\\
BF + LBP$_{P=16}$ &  9.92 & 25.96 & 36 & 1.58 \\
CLBP\_S\_M/C$_{P=16}$ & 11.22 & 27.67 & 54 & 2.30 \\
CLBP\_S/M$_{P=16}$ & 10.23 & 25.97 & 324 & 17.48 \\
\hline
LBP$_{P=24}$ &      7.01 & 17.21 & 26 & 1.11 \\
BF + LBP$_{P=24}$ &      14.95 & 36.69 & 52 & 2.21 \\
CLBP\_S\_M/C$_{P=24}$ & 15.31 & 39.17 & 78 & 4.25 \\
CLBP\_S/M$_{P=24}$ & 14.98 & 37.13 & 676 & 45.63 \\
\hline \hline
\end{tabular}
\end{center}
Time is expressed in milliseconds. With experiments on 1000 images, the average time per image is computed. FET: Feature Extraction Time. FET$_1$ and FET$_2$ are feature extraction time on images of $128\times128$ and $200\times200$ pixels, respectively. MT: Matching Time, this corresponds to the time required for comparing the descriptor of the test image with those of reference images (we consider here a reference set of 2000 samples).\\
\label{tab:time}
\end{table}

\subsubsection{Improving the performance of the CLBC, WLD, and SIFT methods}

\begin{table*}[htbp]
\begin{center}
\caption{CLASSIFICATION RATES OBTAINED ON THE OUTEX DATABASE WHEN USING CLBC AND WLD AS DESCRIPTOR.}
\begin{tabular}{l|rrrr|rrrr|rrrr}
\hline \hline
&\multicolumn{4}{c}{$R=1, P=8$}&\multicolumn{4}{c}{$R=2, P=16$}&\multicolumn{4}{c}{$R=3, P=24$}\\
\hline
& \multicolumn{1}{c}{TC10} & \multicolumn{2}{c}{TC12} & {Average} & \multicolumn{1}{c}{TC10} & \multicolumn{2}{c}{TC12} & {Average} & \multicolumn{1}{c}{TC10} & \multicolumn{2}{c}{TC12} & \multicolumn{1}{c}{Average} \\
& & \multicolumn{1}{c}{t184} & \multicolumn{1}{c}{horizon} & & & \multicolumn{1}{c}{t184} & \multicolumn{1}{c}{horizon} & & & \multicolumn{1}{c}{t184} & \multicolumn{1}{c}{horizon} & \\
\hline
\hline
CLBC\_S  & 82.94 & 65.02 & 63.17 & 70.38 & 88.67 & 82.57 & 77.41 & 82.88 & 91.35 & 83.82 & 82.75 & 85.97\\
BF + CLBC\_S & 96.30 & 92.55 & 92.96 & 93.94 & 98.41 & 95.28 & 95.32 & 96.34 & 98.96 & 95.35 & 94.44 & 96.25 \\
\textit{Gain}& \textit{+13.36} & \textit{+27.53} & \textit{+29.79} & \textbf{\textit{+23.56}} & \textit{+9.74} & \textit{+12.71} & \textit{+17.91} & \textbf{\textit{+13.46}} & \textit{+7.61} & \textit{+11.53} & \textit{+11.69} & \textbf{\textit{+10.28}}\\
\hline
\hline
CLBC\_M      & 78.96 & 53.63 & 58.01 & 63.53 & 92.45 & 70.35 & 72.64 & 78.48 & 91.85 & 72.59 & 74.58 & 79.67\\
BF + CLBC\_M & 90.31 & 79.91 & 83.56 & 84.59 & 97.94 & 87.94 & 90.46 & 92.11 & 96.82 & 88.73 & 91.55 & 92.37\\
\textit{Gain}& \textit{+11.35} & \textit{+26.28} & \textit{+25.55} & \textit{+21.06} & \textit{+5.49} & \textit{+17.59} & \textit{+17.82} & \textit{+13.63} & \textit{+4.97} & \textit{+16.14} & \textit{+16.97} & \textit{+12.70}\\
\hline
\hline
CLBC\_S/M & 95.23 & 82.13 & 83.59 & 86.98 & 98.10 & 89.95 & 90.42 & 92.82 & 98.70 & 91.41 & 90.25 & 93.45\\
BF + CLBC\_S/M  & 96.74 & 94.72 & 92.78 & 94.75 & 99.48 & \textbf{97.89} & \textbf{97.31} & \textbf{98.23} & \textbf{99.40} & 97.57 & \textbf{97.92} & 98.30\\
\textit{Gain} & \textit{+1.51} & \textit{+12.59} & \textit{+9.19} & \textit{+7.77} & \textit{+1.38} & \textit{+7.94} & \textit{+6.89} & \textit{+5.41} & \textit{+0.70} & \textit{+6.16} & \textit{+7.67} &\textit{+4.85}\\
\hline
\hline
CLBC\_S/M/C & 97.16 & 89.79 & 92.92 & 93.29 & 98.54 & 93.26 & 94.07 & 95.29 & 98.78 & 94.00 & 93.24 & 95.67\\
BF + CLBC\_S/M/C & \textbf{97.40} & \textbf{95.09} & \textbf{94.12} & \textbf{95.54} & \textbf{99.51} & 96.41 & 95.83 & 97.25 & 99.32 & \textbf{97.78} & 97.85 & \textbf{98.32}\\
\textit{Gain} & \textit{+0.24} & \textit{+5.30} & \textit{+1.20} & \textit{+2.25} & \textit{+0.97} & \textit{+3.15} & \textit{+1.76} & \textit{+1.96} & \textit{+0.54} & \textit{+3.78} & \textit{+4.61} & \textit{+2.65}\\
\hline
\hline
CLBC\_CLBP \cite{clbc2012tip} & 96.88 & 90.25 & 92.92 & 93.35 & 98.83 & 93.59 & 94.26 & 95.56 & 98.96 & 95.37 & 94.72 & 96.35\\
\hline
\hline
\end{tabular}
\begin{tabular}{l|rrrr}
&\multicolumn{1}{c}{TC10} & \multicolumn{2}{c}{TC12} & {Average}\\
&& t184 & horizon &\\
\hline
\hline
WLD & 77.86 & 58.26 & 54.42 & 63.51\\
BF + WLD & 95.70 & 88.33 & 89.03 & 91.02\\
Gain & +17.84 & +30.07 & +34.61 & +26.51\\
\hline
\hline
SIFT & 83.78 & 74.10 & 75.58 & 77.82\\
BF + SIFT & 94.77 & 87.78 & 88.94 & 90.50 \\
Gain & +10.99 & +13.68 & +13.66 & +12.78 \\
\hline
\hline
\end{tabular}

\end{center}
\label{tab:outex2}
\end{table*}

This section considers the performance of the BF filter when combining with the CLBC, WLD and SIFT descriptors. The experimental results of different methods are reported in Table III, from which we can get similar interesting findings as in the experiments with the completed LBP model:

\begin{itemize}
\item For all considered features or combination strategies with their different parameter settings, using the BF filter as preprocessing results in important gains in performance. For example, when combining our BF filtering with the conventional LBC, with the three parameter configurations, the average improvements are respectively 23.56\%, 13.46\%, and 10.28\%. Similarly, when combining the BF filtering with CLBC\_M, the average improvements with the three parameters are respectively 21.06\%, 13.63\%, and 12.70\%. When combining the BF filter with WLD and SIFT, we obtained important improvements of 26.51\% and 12.78\% respectively.
\item With the same parameters, the \emph{simple} combination ``BF $+$ LBC'' \emph{outperforms all}  ``original'' combination schemes.
\item Also, both the simple combinations ``BF + LBP'' and ``BF + LBC'' are comparable (with $R=3$) or even outperform ``CLBC\_CLBP'' \cite{clbc2012tip} (CLBP\_S/M/C + CLBC\_S/M/C) which combines two completed models of features, the completed LBP and the completed LBC, and use six operators (three for each model).
\end{itemize}

\subsubsection{Further Experiments and Discussions}
\begin{figure*}[htbp]
\centering
\includegraphics[width=0.75\textwidth]{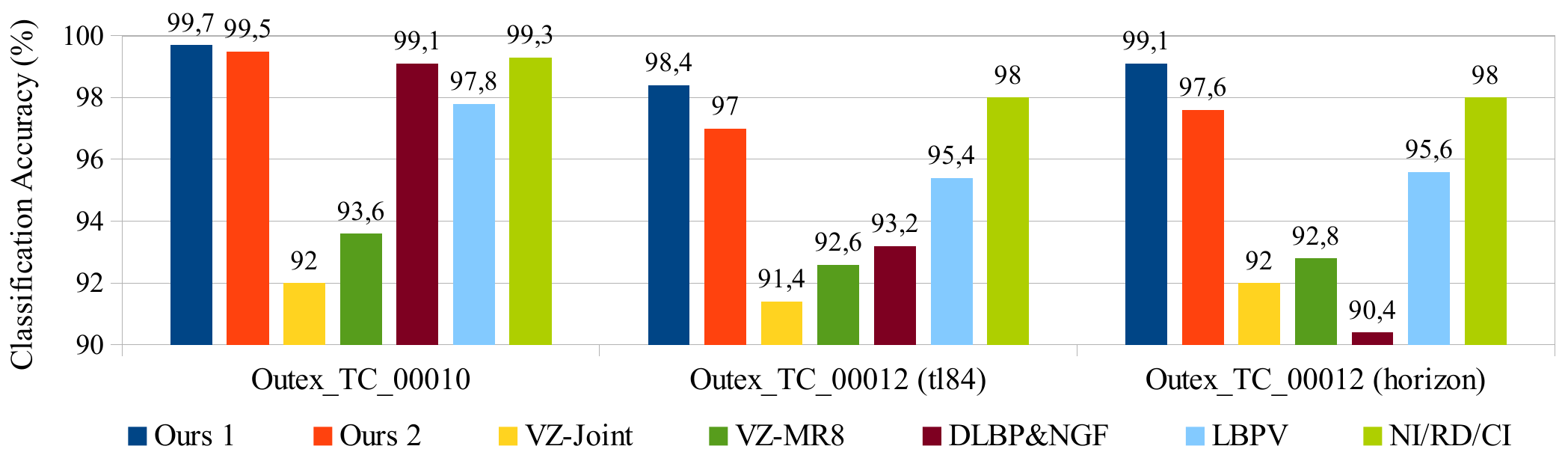}
\caption{Comparison of the best classification scores. Ours 1: BF + CLBP\_S/M $(R=2,P=16)$, Ours 2: \emph{simple} BF + LBP $(R=2,P=16)$. Also, with these optimal parameters, our BF + CLBC\_S/M $(R=2)$ obtains the very high classification rates on these three test suites which are respectively 99.53$\%$, 98.54$\%$, and 98.89$\%$. For a fair comparison, the best results of single scale NI/RD/CI $(R=2)$ \cite{liu2012imavis} are considered in this figure.}
\label{fig:outex3}
\end{figure*}

We compare now our results obtained with the optimal parameters of the BF filter to different state-of-the-art results. From Fig. \ref{fig:outex3}, we can see that:
\begin{itemize}
   \item Our simple `BF $+$ LBP'' already outperforms many state-of-the-art algorithms, including NGF, DLBP \cite{liao2009tip}, and VZ\_MR8, VZ\_Joint \cite{varma2009pami}. The method of Liao et al. in \cite{liao2009tip} has to combine the Normalized Gabor Filter (NGF) and a more complicated LBP variant, DLBP, but its results are still inferior to ours.
  \item Our combination BF + CLBP\_S/M and BF + CLBC\_S/M outperform all the considered algorithms, even the best results of the multi-scale NI/RD/CI (Neighboring Intensities, Radial Difference, and Central Intensity) \cite{liu2012imavis} which are respectively 99.7$\%$, 98.7$\%$, and 98.1$\%$ for the three test suites of the Outex database. To the best of our knowledge, our classification rates obtained are the best ever results on this database.
\end{itemize}

\subsection{Results on the CUReT database}
In the experiments on the CUReT database \footnote{The combination of the BF filter with all CLBP, CLBC, WLD and SIFT descriptors were evaluated on both CURET and UIUC databases and important improvements were obtained. Since the observations are similar, we present in this paper only the results obtained with the CLBP method.}, as in \cite{lazebnik2005pami,clbp2010tip}, to get statistically significant experimental results, $N$ training images were randomly chosen from each class while the remaining $92-N$ images per class were used as the test set (note that, we return to the default parameters of the BF filter).

The average classification rates (with different parameters of CLBP) over a hundred random splits are reported in Table IV, from which we can get similar interesting observations as in the experiments on the Outex database:
\begin{table*}[htbp]
\begin{center}
\caption{CLASSIFICATION RATES OBTAINED ON THE CURET DATABASE.}
\begin{tabular}{l|rrrr|rrrr|rrrr}
\hline \hline
&\multicolumn{4}{c}{$R=1, P=8$}&\multicolumn{4}{c}{$R=3, P=16$}&\multicolumn{4}{c}{$R=5, P=24$}\\
\hline
N & \multicolumn{1}{c}{46} & \multicolumn{1}{c}{23} & \multicolumn{1}{c}{12} & \multicolumn{1}{c}{6} & \multicolumn{1}{c}{46} & \multicolumn{1}{c}{23} & \multicolumn{1}{c}{12} & \multicolumn{1}{c}{6} & \multicolumn{1}{c}{46} & \multicolumn{1}{c}{23} & \multicolumn{1}{c}{12} & \multicolumn{1}{c}{6}\\
\hline
CLBP\_S      & 80.63 & 74.81 & 67.84 & 58.70 & 86.37 & 81.05 & 74.62 & 66.17 & 86.37 & 81.21 & 74.71 & 66.55\\
BF + CLBP\_S & 91.97 & 86.43 & 81.08 & 72.17 & 93.16 & 88.13 & 81.94 & 74.45 & 89.19 & 86.22 & 77.91 & 69.67\\
\textit{Gain} &  \textit{+11.34} &  \textit{+11.62} &  \textit{+13.24} &  \textit{+13.47} &  \textit{+6.79} &  \textit{+7.08} &  \textit{+7.32} &  \textit{+8.28} &  \textit{+2.82} &  \textit{+5.01} &  \textit{+3.20} &  \textit{+3.12}\\
 \hline
\hline
CLBP\_M      & 75.20 & 67.96 & 60.27 & 51.49 & 85.48 & 79.01 & 71.24 & 61.59 & 82.16 & 76.23 & 69.22 & 60.45\\
BF + CLBP\_M & 91.53 & 85.81 & 78.04 & 70.18 & 92.78 & 89.43 & 82.74 & 74.82 & 92.43 & 87.57 & 81.65 & 74.01\\
\textit{Gain} &  \textit{+16.33} &  \textit{+17.85} &  \textit{+17.77} &  \textit{+18.69} &  \textit{+7.30} &  \textit{+10.42} &  \textit{+11.50} &  \textit{+13.23} &  \textit{+10.27} &  \textit{+11.34} &  \textit{+12.43} &  \textit{+13.56}\\
\hline
\hline
CLBP\_M/C      & 83.26 & 75.58 & 66.91 & 56.45 & 91.42 & 85.73 & 78.05 & 68.14 & 89.48 & 83.54 & 75.96 & 66.41\\
BF + CLBP\_M/C & 94.63 & 90.84 & 83.87 & 76.45 & 95.64 & 92.40 & 85.67 & 78.76 & 94.50 & 89.67 & 83.98 & 76.68\\
\textit{Gain} &  \textit{+11.37} &  \textit{+15.26} &  \textit{+16.96} &  \textit{+20.00} &  \textit{+4.22} &  \textit{+6.67} &  \textit{+7.62} &  \textit{+10.62} &  \textit{+5.02} &  \textit{+6.13} &  \textit{+8.02} &  \textit{+10.27}\\
\hline
\hline
CLBP\_S\_M/C&  90.34 & 84.52 & 76.42 & 66.31 & 93.87 & 89.05 & 82.46 & 72.51 & 93.22 & 88.37 & 81.44 & 72.01\\
BF + CLBP\_S\_M/C & 95.68 & 91.77 & 86.77 & 78.97 & 96.08 & 92.46 & 85.28 & 80.84 & 95.01 & 91.99 & 84.52 & 77.67 \\
\textit{Gain} &  \textit{+5.34} &  \textit{+7.25} &  \textit{+10.35} &  \textit{+12.66} &  \textit{+2.21} &  \textit{+3.41} &  \textit{+2.82} &  \textit{+8.33} &  \textit{+1.79} &  \textit{+3.62} &  \textit{+3.08} &  \textit{+5.66}\\
\hline
\hline

CLBP\_S/M &      93.52 & 88.67 & 81.95 & 72.30 & 94.45 & 90.40 & 84.17 & 75.39 & 93.63 & 89.14 & 82.47 & 73.26\\
BF + CLBP\_S/M & 96.76 & 93.85 & 88.02 & 82.97 & \textbf{97.65} & \textbf{94.91} & \textbf{89.13} & \textbf{82.79} & 97.21 & 92.49 & 88.37 & 81.68\\
   
\textit{Gain} &  \textit{+3.24} &  \textit{+5.18} &  \textit{+6.07} &  \textit{+10.67} &  \textit{+3.20} &  \textit{+4.51} &  \textit{+4.96} &  \textit{+7.40} &  \textit{+3.58} &  \textit{+3.35} &  \textit{+5.90} &  \textit{+8.42}\\
\hline
\hline
CLBP\_S/M/C &      95.59 & 91.35 & 84.92 & 74.80 & 95.86 & 92.13 & 86.15 & 77.04 & 94.74 & 90.33 & 83.82 & 74.46\\
BF + CLBP\_S/M/C & 97.20 & 93.89 & 89.26 & 82.75 & 97.04 & 94.25 & 88.57 & 82.02 & 95.82 & 91.69 & 86.97 & 79.24\\
\textit{Gain} &  \textit{+1.61} & \textit{+2.54} & \textit{+4.34} & \textit{+7.95} & \textit{+1.18} & \textit{+2.12} & \textit{+2.42} & \textit{+4.98} & \textit{+1.08} & \textit{+1.36} & \textit{+3.15} & \textit{+4.78}\\
 
\hline
\hline
VZ\_MR8 & \multicolumn{12}{c}{97.79 (46), 95.03 (23), 90.48 (12), 82.90 (6)}\\
VZ\_Joint & \multicolumn{12}{c}{97.66 (46), 94.58 (23), 89.40 (12), 81.06 (6)}\\
Multiscale NI/RD/CI & \multicolumn{12}{c}{97.29 (46)}\\
\hline
\hline
\end{tabular}
\end{center}
\label{tab:curet}
\end{table*}

\begin{itemize}
\item For all considered methods with different parameters, using our algorithm as preprocessing results in
important gains in performance. For example, when combining BF with CLBP\_M, with (R=1, P=8), the improvements
with four different numbers of training images used ($N=46, 23, 12, 6$) are respectively 16.33$\%$, 17.85$\%$,
17.77$\%$, and 18.69$\%$.

\item In the CUReT database, there are scale and affine variations. While VZ\_MR8 and VZ\_Joint were
designed with scale and affine invariance property, the CLBP operators we used have limited capability
to address scale and affine invariance. However, interestingly, ``BF + CLBP\_S/M/C'' still performs
as well as VZ\_MR8 and VZ\_Joint. It also outperforms the recent multi-scale NI/RD/CI method \cite{liu2012imavis} which must combine three descriptors at three scales.
\end{itemize}

\subsection{Results on the UIUC database}

\begin{table}[htbp]
\begin{center}
\caption{CLASSIFICATION RATES OBTAINED ON THE UIUC DATABASE.}
\begin{tabular}{l|l|rrrr}
\hline \hline
& & \multicolumn{1}{c}{20} & \multicolumn{1}{c}{15} & \multicolumn{1}{c}{10} & \multicolumn{1}{c}{5}\\
\hline
\multirow{12}{*}{\rotatebox{90}{\mbox{R=1, P=8}}} 
& CLBP\_S & 54.78 & 51.85 & 46.79 & 40.53\\
& BF + CLBP\_S & 87.79 & 77.24 & 75.82 & 69.13\\
& \textit{Gain} & \textit{+33.01} & \textit{+25.39} & \textit{+29.03} & \textit{+28.60}\\

\cline{2-6}
& CLBP\_M & 57.52 & 54.14 & 50.11 & 40.95\\
& BF + CLBP\_M & 83.63 & 75.26 & 72.04 & 63.11\\
& \textit{Gain} & \textit{+26.11} & \textit{+21.12} & \textit{+21.93} & \textit{+22.16}\\
\cline{2-6}
& CLBP\_S/M & 81.80 & 78.55 & 74.8 & 64.84\\
& BF + CLBP\_S/M & 91.21 & 87.15 & 83.19 & 75.65\\
& \textit{Gain} & \textit{+9.41} & \textit{+8.60} & \textit{+8.39} & \textit{+10.81}\\

\cline{2-6}
& CLBP\_S/M/C & 87.64 & 85.70 & 82.65 & 75.05 \\
& BF + CLBP\_S/M/C & \textbf{91.47} & \textbf{87.41} & \textbf{83.96} & \textbf{76.24}\\
& \textit{Gain} & \textit{+3.83} & \textit{+1.71} & \textit{+1.31} & \textit{+1.19}\\
 
\hline \hline
\multirow{12}{*}{\rotatebox{90}{\mbox{R=2, P=16}}}
& CLBP\_S & 61.04 & 55.84 & 51.77 & 41.88\\
& BF + CLBP\_S & 90.76 & 88.13 & 85.07 & 76.22\\
&\textit{Gain} & \textit{+29.72} & \textit{+32.29} & \textit{+33.30} & \textit{+34.34}\\
\cline{2-6}
& CLBP\_M & 72.12 & 68.99 & 64.47 & 57.06\\
& BF + CLBP\_M & 87.98 & 84.37 & 81.76 & 70.95\\
& \textit{Gain} & \textit{+15.86} & \textit{+15.38} & \textit{+17.29} & \textit{+13.89}\\
\cline{2-6}
& CLBP\_S/M & 87.87 & 85.07 & 80.59 & 71.64\\
& BF + CLBP\_S/M & \textbf{94.25} & \textbf{91.77} & \textbf{89.68} & \textbf{80.86}\\
& \textit{Gain} & \textit{+6.38} & \textit{+6.70} & \textit{+9.09} & \textit{+9.22} \\
\cline{2-6}
& CLBP\_S/M/C & 91.04 & 89.42 & 86.29 & 78.57\\
& BF + CLBP\_S/M/C & 93.45 & 90.99 & 87.97 & 80.19\\
& \textit{Gain} & \textit{+2.41} & \textit{+1.57} & \textit{+1.68} & \textit{+1.62} \\
\hline
\hline

\multirow{12}{*}{\rotatebox{90}{\mbox{R=3, P=16}}}
& CLBP\_S & 64.11 & 60.11 & 54.67 & 44.45\\
& BF + CLBP\_S & 91.24 & 87.46 & 82.87 & 73.12\\
&\textit{Gain} & \textit{+27.13} & \textit{+27.35} & \textit{+28.20} & \textit{+28.67}\\
\cline{2-6}
& CLBP\_M & 74.45 & 71.47 & 65.21 & 56.72\\
& BF + CLBP\_M & 89.78 & 86.01 & 81.07 & 71.12\\
& \textit{Gain} & \textit{+15.33} & \textit{+14.54} & \textit{+15.86} & \textit{+14.44}\\
\cline{2-6}
& CLBP\_S/M & 89.18 & 87.42 & 81.95 & 72.53\\
& BF + CLBP\_S/M & \textbf{94.14} & \textbf{91.99} & \textbf{89.61} & \textbf{80.98}\\
& \textit{Gain} & \textit{+4.96} & \textit{+4.57} & \textit{+7.66} & \textit{+8.45}\\

& CLBP\_S/M/C & 91.19 & 89.21 & 85.95 & 78.05\\
& BF + CLBP\_S/M/C & 93.78 & 91.64 & 88.12 & 80.23\\
& \textit{Gain} & \textit{+2.59} & \textit{+2.43} & \textit{+2.17} & \textit{+2.18}\\
 
\hline
\hline
\multicolumn{2}{l}{Multi-scale CLBP\_S/M/C:} & 91.57 & 89.84 & 86.73 & 78.42\\
\multicolumn{2}{l}{Multi-scale CLBC\_S/M/C:} & 92.42 & 90.66 & 87.75 & 80.22\\
\hline
\hline
\end{tabular}
\end{center}
\centering {Multi-scale CLBP\_S/M/C and Multi-scale CLBC\_S/M/C must combine all the three scales $R = 1, 2, 3$.}
\end{table}


As in \cite{lazebnik2005pami}, to eliminate the dependence of the results on the particular training
images used, $N$ training images were randomly chosen from each class while the remaining $40-N$ images
per class were used as test set.

The average classification rates over a hundred random splits are reported in Table V.
Consistent with the analysis of the results obtained on two other databases, we have many
interesting observations, and we highlight here two observations:
\begin{itemize}
\item For all considered methods with different parameters, using our algorithm as preprocessing
results in important gains in performance. For example, when combining BF with CLBP\_S, with
(R=1, P=8), the improvements with four different numbers of training images used ($N=20, 15, 10, 5$)
are respectively 31.51$\%$, 26.71$\%$, 28.28$\%$, and 28.12$\%$.
\item Also, the combination ``BF + \emph{single-scale} CLBP\_S/M/C'' (with $R=2, 3$) outperforms
both multi-scale CLBP\_S/M/C and multi-scale CLBC\_S/M/C which must combine all the three scales $R = 1, 2, 3$.
\end{itemize}

\subsection{Robustness to Noise}

Robustness to noise is one of the most important factors to assess texture classification methods. To measure the robustness of the proposed method we use the three test suites of the Outex dataset. In our experiments, the original texture images are added with random Gaussian noise with different signal-to-noise ratios
(SNR). In particular, we study the robustness of the methods with six levels of noise SNR = \{30, 15, 10, 5, 4, 3\}. To reduce the variability of randomness, each experiment is repeated ten times and then the average classification accuracies and the standard deviations are calculated for each test suite and for each descriptor or combination strategy.

As can be seen from Table VI, the proposed method is very robust to noise. When the proposed preprocessing is used in addition, the classification accuracy is much more stable for all evaluated descriptors. One can see that the WLD descriptor is very sensitive to noise and its performance drops quickly when the SNR is smaller than 15. The performance of the combination BF and SIFT drops more than other combinations (about 13\% with SNR=3) since when using SIFT, the additional noise influences not only the feature extraction but also the keypoint detection step (the preprocessing method presented in this paper is related to only the feature extraction). The other considered descriptors keep their performance for SNR=10, however, their performance drops suddenly when the SNR is decreased to 5. This loss is about 5\% for the combination strategies such as CLBP\_S/M and CLBC\_S/M, and more than 12\% for the ``individual'' descriptor like LBC and LBP. When the BF filter is applied, for CLBP\_S/M and CLBC\_S/M, the losses are around 0.5\% with SNR=5 for all three test suites. When combining the BF filter with `individual'' descriptors like LBP and LBC, the losses are about 1.5\% with SNR=5.

\begin{table*}[htbp]
\begin{center}
\caption{CLASSIFICATION RATES ON THE OUTEX TEST SUITES WITH DIFFERENT SIGNAL-TO-NOISE.}
\begin{tabular}{l|l|cccccc}
\hline \hline
&& SNR=30 & SNR=15 & SNR=10 & SNR=5 & SNR=4 & SNR=3\\

\hline\multirow{10}{*}{\rotatebox{90}{\mbox{TC10}}}

& CLBP\_S/M$_{R=2}$ & 97.87 $\pm$ 0.15 & 97.45 $\pm$ 0.29 & 96.70 $\pm$ 0.42 & 91.01
$\pm$ 0.68 & 88.99 $\pm$ 0.87 & 86.81 $\pm$ 1.17\\
& BF + CLBP\_S/M$_{R=2}$ &  99.26 $\pm$ 0.06 & 99.27 $\pm$ 0.07 & 99.17 $\pm$ 0.18 & 98.76
$\pm$ 0.34 & 98.52 $\pm$ 0.41 & 98.04 $\pm$ 0.45\\

\cline{2-8}
& LBC$_{R=2}$ & 88.61 $\pm$ 0.18 &  88.45  $\pm$  0.29  &  87.16  $\pm$  0.88  &  75.31
 $\pm$   1.02 &  69.07  $\pm$  1.45 &  65.68  $\pm$  1.69\\
& BF + LBC$_{R=2}$  &  98.38 $\pm$   0.07 &  98.37 $\pm$ 0.08 &  98.03  $\pm$  0.19 &  96.87 $\pm$
 0.30 &  95.57 $\pm$ 0.41 &  95.02 $\pm$ 0.72\\

\cline{2-8}
& CLBC\_S/M$_{R=2}$ & 97.91 $\pm$ 0.18 & 97.59 $\pm$ 0.32 & 96.83 $\pm$ 0.49 & 91.23
$\pm$ 0.67 & 89.01 $\pm$ 0.89 & 86.97 $\pm$ 1.13\\
& BF + CLBC\_S/M$_{R=2}$ &  99.46 $\pm$ 0.07 & 99.47 $\pm$ 0.08 & 99.28 $\pm$ 0.16 & 98.89
$\pm$ 0.33 & 98.76 $\pm$ 0.40 & 98.27 $\pm$ 0.44\\

\cline{2-8}
& WLD      & 84.61 $\pm$ 0.41 & 80.51 $\pm$ 0.42 & 70.78 $\pm$ 0.79 & 48.64 $\pm$ 1.28 & 40.46 $\pm$ 2.08 & 34.51 $\pm$ 2.75\\
& BF + WLD & 95.79 $\pm$ 0.17 & 95.72 $\pm$ 0.20 & 95.66 $\pm$ 0.31 & 94.84 $\pm$ 0.42 & 93.49 $\pm$ 0.69 & 92.74 $\pm$ 0.93\\

\cline{2-8}
& SIFT      & 82.23 $\pm$ 0.47 & 80.45 $\pm$ 0.44 & 76.12 $\pm$ 0.82 & 68.61 $\pm$ 1.32 & 65.36 $\pm$ 1.91 & 63.21 $\pm$ 2.56\\
& BF + SIFT & 94.29 $\pm$ 0.19 & 93.18 $\pm$ 0.23 & 91.45 $\pm$ 0.34 & 85.04 $\pm$ 0.49 & 83.94 $\pm$ 0.78 & 81.62 $\pm$ 0.96\\

\hline\hline
\multirow{10}{*}{\rotatebox{90}{\mbox{TC12t}}}
& CLBP\_S/M$_{R=2}$ & 90.44 $\pm$ 0.22 & 89.71 $\pm$ 0.33 & 88.01 $\pm$ 0.52 & 82.61
$\pm$ 0.68 & 80.02 $\pm$ 0.93 & 78.38 $\pm$ 1.01\\
& BF + CLBP\_S/M$_{R=2}$ &  98.16 $\pm$ 0.06 & 98.07 $\pm$ 0.09 & 97.82 $\pm$ 0.23 & 97.08
$\pm$ 0.37 & 96.95 $\pm$ 0.41 & 96.44 $\pm$ 0.49\\

\cline{2-8}
& LBC$_{R=2}$ & 82.11 $\pm$ 0.19 &  82.35  $\pm$  0.28  &  80.36  $\pm$  1.25  &  67.17
 $\pm$   1.40 &  62.06  $\pm$  1.67 &  57.75  $\pm$  2.41\\
& BF + LBC$_{R=2}$  &  95.08 $\pm$   0.09 &  94.67 $\pm$ 0.17 &  94.13  $\pm$  0.30 &  91.75 $\pm$
 0.33 &  91.01 $\pm$ 0.42 &  90.52 $\pm$ 0.92\\

\cline{2-8}
& CLBC\_S/M$_{R=2}$ & 89.85 $\pm$ 0.21 & 88.59 $\pm$ 0.37 & 86.93 $\pm$ 0.53 & 81.85
$\pm$ 0.71 & 79.38 $\pm$ 0.97 & 77.97 $\pm$ 1.09\\
& BF + CLBC\_S/M$_{R=2}$ &  97.96 $\pm$ 0.08 & 97.92 $\pm$ 0.11 & 97.78 $\pm$ 0.21 & 97.19
$\pm$ 0.38 & 96.99 $\pm$ 0.43 & 96.50 $\pm$ 0.48\\

\cline{2-8}
& CLBC\_S/M/C$_{R=2}$ &  94.05 $\pm$ 0.08 & 93.62 $\pm$ 0.15 & 92.53 $\pm$ 0.21 & 89.39
$\pm$ 0.38 & 87.81 $\pm$ 0.43 & 85.96 $\pm$ 0.48\\
& BF + CLBC\_S/M/C$_{R=2}$ &  96.85 $\pm$ 0.08 & 96.62 $\pm$ 0.15 & 96.78 $\pm$ 0.21 & 96.39
$\pm$ 0.38 & 96.08 $\pm$ 0.43 & 95.69 $\pm$ 0.48\\

\cline{2-8}
&WLD      & 60.61 $\pm$ 0.54 & 59.29 $\pm$ 0.63 & 53.80 $\pm$ 0.73 & 38.14 $\pm$ 1.41 & 34.62 $\pm$ 2.37 & 29.44 $\pm$ 2.93\\
&BF + WLD & 87.71 $\pm$ 0.16 & 87.39 $\pm$ 0.39 & 86.91 $\pm$ 0.52 & 85.86 $\pm$ 0.75 & 84.64 $\pm$ 0.86 & 83.45
$\pm$ 1.02\\

\hline\hline
\multirow{10}{*}{\rotatebox{90}{\mbox{TC12h}}}
& CLBP\_S/M$_{R=2}$ & 91.07 $\pm$ 0.16 & 90.51 $\pm$  0.27 &  88.41  $\pm$  0.44 & 82.71
 $\pm$ 0.66 &  81.02  $\pm$  0.89 &  78.03 $\pm$ 0.95\\
& BF + CLBP\_S/M$_{R=2}$ & 97.75 $\pm$ 0.07  & 97.64 $\pm$ 0.09 & 97.51 $\pm$ 0.18 & 97.27 $\pm$
   0.27 &  96.99 $\pm$  0.33 &  96.21 $\pm$  0.46\\

\cline{2-8}
& LBC$_{R=2}$ & 76.93 $\pm$ 0.39 & 76.81 $\pm$ 0.58 & 73.98 $\pm$ 0.93 & 61.03 $\pm$ 1.10 & 58.72  $\pm$ 1.98 & 54.51 $\pm$ 2.04\\
& BF + LBC$_{R=2}$ & 95.26 $\pm$ 0.10 & 94.93 $\pm$ 0.29 & 94.65 $\pm$ 0.37 & 93.78
$\pm$ 0.59 & 92.69 $\pm$ 0.78 & 91.58 $\pm$ 0.89\\

\cline{2-8}
& CLBC\_S/M$_{R=2}$ & 90.24 $\pm$ 0.19 & 89.57 $\pm$  0.29 &  88.55  $\pm$  0.46 & 81.82
 $\pm$ 0.69 &  80.20  $\pm$  0.87 &  77.19 $\pm$ 0.98\\
& BF + CLBC\_S/M$_{R=2}$ & 97.54 $\pm$ 0.07  & 97.45 $\pm$ 0.10 & 97.35 $\pm$ 0.19 & 96.74 $\pm$
   0.28 &  96.58 $\pm$  0.32 &  95.96 $\pm$  0.45\\

\cline{2-8}
& CLBC\_S/M/C$_{R=2}$ & 94.30 $\pm$  0.21 &  93.88 $\pm$  0.43  & 93.21  $\pm$  0.73 & 90.44 $\pm$
    0.96 &  89.04  $\pm$  1.22 &  88.01 $\pm$   1.58\\
& BF + CLBC\_S/M/C$_{R=2}$ &   96.69 $\pm$   0.12  & 96.45  $\pm$  0.18  & 96.07 $\pm$  0.38  & 95.73
 $\pm$  0.52 &  95.31  $\pm$  0.63 &  94.67  $\pm$  0.69\\

\cline{2-8}
&WLD      & 65.29 $\pm$ 0.57 & 63.24 $\pm$ 0.79 & 57.94 $\pm$ 0.95 & 38.69 $\pm$ 1.23 & 35.47 $\pm$ 2.12 & 29.93 $\pm$ 2.87 \\
&BF + WLD & 88.84 $\pm$ 0.21 & 88.29 $\pm$ 0.35 & 87.94 $\pm$ 0.46 & 87.62 $\pm$ 0.59 & 86.41 $\pm$ 0.71 & 86.12 $\pm$ 0.94 \\
\hline \hline

\end{tabular}
\end{center}
\label{tab:noise}
\end{table*}
\section{Conclusion}\label{ss:conclusion}
With the objective of improving texture classification systems at the level of preprocessing, a novel,
simple, yet very powerful biologically-inspired algorithm simulating the performance of human retina
has been described. After applying a DoG filter to detect the edges, the filtered image is first split
into two ``maps'' alongside the sides of its edges, and the feature extraction step is then carried out
on these two ``maps'' instead of the input image. Experiment results on large databases validate the
efficiency of the proposed method both in terms of high performance and low complexity. The proposed algorithm was also proved to be robust to noise. Future work involves evaluating in more depth the performance of the proposed method for other texture classification methods  as well as other pattern recognition tasks such as face recognition.



\end{document}